\documentclass[letterpaper,twocolumn,10pt]{article}
\usepackage{usenix2019_v3}

\usepackage{tikz}
\usepackage{amsmath}
\usepackage{xspace}
\usepackage{multirow}
\usepackage{multicol}
\usepackage{ifthen}
\usepackage{subfig}
\usepackage{caption}
\usepackage{enumitem}
\usepackage[available, functional, reproduced]{usenixbadges}

\newcommand{\Ansor}{Ansor\xspace}

\newcommand{\showcomments}{yes}

\newcommand\todo[1]{
    \ifthenelse{\equal{\showcomments}{yes}}{{\color{red} TODO: #1}}{\ignorespaces}
}
\newcommand\joey[1]{
    \ifthenelse{\equal{\showcomments}{yes}}{{\color{cyan} (Joey: #1)}}{\ignorespaces}
}
\newcommand\lianmin[1]{
    \ifthenelse{\equal{\showcomments}{yes}}{{\color{blue} Lianmin: #1}}{\ignorespaces}
}
\newcommand\danyang[1]{
    \ifthenelse{\equal{\showcomments}{yes}}{{\color{blue} Danyang: #1}}{\ignorespaces}
}
\newcommand\ion[1]{
    \ifthenelse{\equal{\showcomments}{yes}}{{\color{blue} Ion: #1}}{\ignorespaces}
}
\newcommand\ks[1]{
    \ifthenelse{\equal{\showcomments}{yes}}{{\color{blue} Koushik: #1}}{\ignorespaces}
}
\newcommand\yida[1]{
    \ifthenelse{\equal{\showcomments}{yes}}{{\color{blue} [Yida: #1]}}{\ignorespaces}
}
\newcommand\cody[1]{
    \ifthenelse{\equal{\showcomments}{yes}}{{\color{blue} [Cody: #1]}}{\ignorespaces}
}
\newcommand\ameer[1]{
    \ifthenelse{\equal{\showcomments}{yes}}{{\color{blue} Ameer: #1}}{\ignorespaces}
}
\newcommand\lsf[1]{
    \ifthenelse{\equal{\showcomments}{yes}}{{\color{blue} Frank: #1}}{\ignorespaces}
}

\newcommand{\preeq}{\vskip -0.5em}
\newcommand{\posteq}{\vskip -0.5em}

\newcommand{\precap}{\vskip -1.0em}
\newcommand{\postcap}{\vskip -1.2em}

\newcommand{\postsec}{\vskip -0.8em}

\def\Snospace~{\S{}}

\newcommand{\autorefsuffix}[2]{\hyperref[#1]{\autoref*{#1}#2}}

\begin{document}

\date{}

\title{ \Ansor: Generating High-Performance Tensor Programs for Deep Learning}

\author{
\rm{Lianmin Zheng $ ^1$,
  Chengfan Jia $ ^2$,
  Minmin Sun $ ^2$,
 Zhao Wu $ ^2$,
Cody Hao Yu $ ^3$} ,\\
\rm{   Ameer Haj-Ali $ ^1$,
  Yida Wang $ ^3$,
  Jun Yang $ ^2$,
  Danyang Zhuo $ ^{1,4}$ }, \\
\rm{Koushik Sen $ ^1$,
	  Joseph E. Gonzalez $ ^1$,
  Ion Stoica $ ^1$}\\
\\
  {$ ^1$ UC Berkeley,\ $ ^2$Alibaba Group,\ $ ^3$Amazon Web Services, $ ^4$ Duke University}
} %

\maketitle
\begin{abstract}

High-performance tensor programs are crucial to guarantee efficient execution of deep neural networks. However, obtaining performant tensor programs for different operators on various hardware platforms is notoriously challenging. Currently, deep learning systems rely on vendor-provided kernel libraries or various search strategies to get performant tensor programs.
These approaches either require significant engineering effort to develop platform-specific optimization code or fall short of finding high-performance programs due to restricted search space and ineffective exploration strategy.

We present \Ansor, a tensor program generation framework for deep learning applications. 
Compared with existing search strategies, \Ansor explores many more optimization combinations by sampling programs from a hierarchical representation of the search space. 
\Ansor then fine-tunes the sampled programs with evolutionary search and a learned cost model to identify the best programs.
\Ansor can find high-performance programs that are outside the search space of existing state-of-the-art approaches.
In addition, \Ansor utilizes a task scheduler to simultaneously optimize multiple subgraphs in deep neural networks.
We show that \Ansor improves the execution performance of deep neural networks relative to the state-of-the-art on the Intel CPU, ARM CPU, and NVIDIA GPU by up to  $3.8\times$, $2.6\times$, and $1.7\times$, respectively.

\end{abstract}
\section{Introduction}

Low-latency execution of deep neural networks (DNN) plays a critical role in autonomous driving \cite{cordts2016cityscapes}, augmented reality \cite{alhaija2017augmented}, language translation\cite{devlin2018bert}, and other applications of AI.
DNNs can be expressed as a directed acyclic computational graph (DAG), in which nodes represent the operators (\textit{e.g.}, convolution, matrix multiplication) and directed edges represent the dependencies between operators.
Existing deep learning frameworks (\textit{e.g.}, Tensorflow~\cite{abadi2016tensorflow}, PyTorch~\cite{paszke2019pytorch}, MXNet~\cite{chen2015mxnet}) map the operators in DNNs to vendor-provided kernel libraries (\textit{e.g.}, cuDNN~\cite{chetlur2014cudnn}, MKL-DNN~\cite{intel2017mkldnn}) to achieve high performance.
However, these kernel libraries require significant engineering effort to manually tune for each hardware platform and operator.
The significant manual effort required to produce efficient operator implementations for each target accelerator limits the development and innovation of new operators~\cite{barham2019machine} and specialized accelerators~\cite{moreau2019hardware}.

Given the importance of DNNs' performance, researchers and industry practitioners have turned to search-based compilation~\cite{chen2018tvm, adams2019learning,zheng2020flextensor,vasilache2018tensor,liu2019optimizing} for automated generation of \emph{tensor programs}, \textit{i.e.}, low-level implementations of tensor operators. 
For an operator or a (sub-)graph of multiple operators, users define the computation in a high-level declarative language (\autoref{sec:background}), and the compiler then searches for programs tailored towards different hardware platforms.

To find performant tensor programs, it is necessary for a search-based approach to explore a large enough search space to cover all the useful tensor program optimizations. 
However, existing approaches fail to capture many effective optimization combinations, because they rely on either predefined manually-written templates (\textit{e.g.}, TVM~\cite{chen2018learning}, FlexTensor~\cite{zheng2020flextensor}) or aggressive pruning by evaluating incomplete programs (\textit{e.g.}, Halide auto-scheduler \cite{adams2019learning}), which prevents them from covering a comprehensive search space (\autoref{sec:background}).
The rules they use to construct the search space are also limited.

In this paper, we explore a novel search strategy for generating high-performance tensor programs.
It can automatically generate a large search space with comprehensive coverage of optimizations and gives every tensor program in the space a chance to be chosen. 
It thus enables to find high-performance programs that existing approaches miss.

Realizing this goal faces multiple challenges.
First, it requires automatically constructing a large search space to cover as many tensor programs as possible for a given computation definition.
Second, we need to search efficiently without comparing incomplete programs in the large search space that can be orders of magnitude larger than what existing templates can cover.
Finally, when optimizing an entire DNN with many subgraphs, we should recognize and prioritize the subgraphs that are critical to the end-to-end performance.

To this end, we design and implement \emph{\Ansor}, a framework for automated tensor program generation.
\Ansor utilizes a hierarchical representation to cover a large search space.
This representation decouples high-level structures and low-level details, enabling flexible enumeration of high-level structures and efficient sampling of low-level details.
The space is constructed automatically for a given computation definition.
\Ansor then samples complete programs from the search space and fine-tunes these programs with evolutionary search and a learned cost model.
To optimize the performance of DNNs with multiple subgraphs, 
\Ansor dynamically prioritizes subgraphs of the DNNs that are more likely to improve the end-to-end performance.

We evaluate \Ansor on both standard deep learning benchmarks and emerging new workloads against manual libraries and state-of-the-art search-based frameworks.
Experiment results show that \Ansor improves the execution performance of DNNs on the Intel CPU, ARM CPU, and NVIDIA GPU by up to  $3.8\times$, $2.6\times$, and $1.7\times$, respectively.
For most computation definitions, the best program found by \Ansor is outside the search space of existing search-based approaches.
The results also show that, compared with existing search-based approaches, \Ansor searches more efficiently, generating higher-performance programs in a shorter time, despite its larger search space.
\Ansor can match the performance of a state-of-the-art framework with an order of magnitude less search time.
Besides, \Ansor enables automatic extension to new operators by only requiring their mathematical definitions without manual templates.

In summary, this paper makes the following contributions:

\begin{itemize}[topsep=-0.1em, itemsep=-0.1em]
    \item A mechanism to generate a large hierarchical search space of tensor programs for a computational graph.
    \item An evolutionary strategy with a learned cost model to fine-tune the performance of tensor programs.
    \item A scheduling algorithm based on gradient descent to prioritize important subgraphs when optimizing the end-to-end performance of DNNs.
    \item An implementation and comprehensive evaluation of the \Ansor system demonstrating that the above techniques outperform state-of-the-art systems on a variety of DNNs and hardware platforms.
\end{itemize}

\section{Background}
\label{sec:background}

The deep learning ecosystem is embracing a rapidly growing diversity of hardware platforms including CPUs, GPUs, FPGAs, and ASICs.
In order to deploy DNNs on these platforms, high-performance tensor programs are needed for the operators used in DNNs.
The required operator set typically contains a mixture of standard operators (\textit{e.g.}, matmul, conv2d) and novel operators invented by machine learning researchers (\textit{e.g.}, capsule conv2d~\cite{hinton2018matrix}, dilated conv2d~\cite{yu2015multi}).

\begin{figure}[t]
	\centering
	\vskip -1.0em
	\includegraphics[width=1.01\columnwidth]{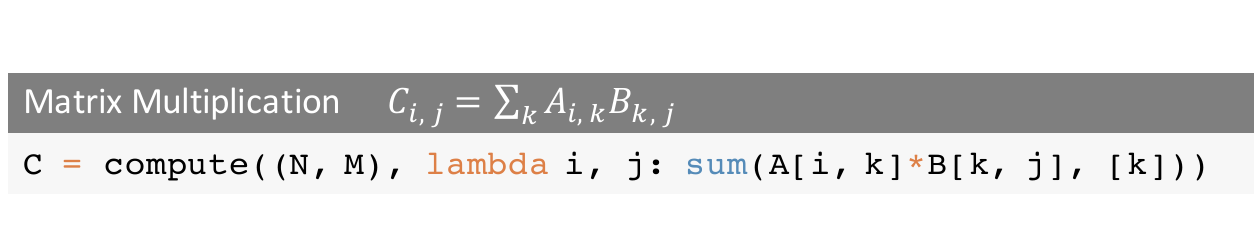}
	\precap 
	\caption{The computation definition of matrix multiplication.}
	\postcap \vskip -0.5em
	\label{fig:compute-definition-examples}
\end{figure}

To deliver portable performance of these operators on a wide range of hardware platforms in a productive way, multiple
compiler techniques have been introduced (\textit{e.g.},  TVM~\cite{chen2018tvm}, Halide~\cite{ragan2013halide}, Tensor Comprehensions~\cite{vasilache2018tensor}).
Users define the computation in a form similar to mathematical expressions using a high-level declarative language, and the compiler generates optimized tensor programs according to the definition.
\autoref{fig:compute-definition-examples} shows the computation definition of matrix multiplication in the TVM tensor expression language. Users mainly need to define the shapes of the tensors and how each element in the output tensor is computed.

However, automatically generating high-performance tensor programs from a high-level definition is extremely difficult.
Depending on the architecture of the target platform, the compiler needs to search in an extremely large and complicated space containing combinatorial choices of optimizations (\textit{e.g.}, tile structure, tile size, vectorization, parallelization).
Finding high-performance programs requires the search strategy to cover a comprehensive space and explore it efficiently.
We describe two recent and effective approaches in this section and other related work in \autoref{sec:related-work}.

\begin{figure*}[t!]
	\centering
	\includegraphics[width=0.95\textwidth]{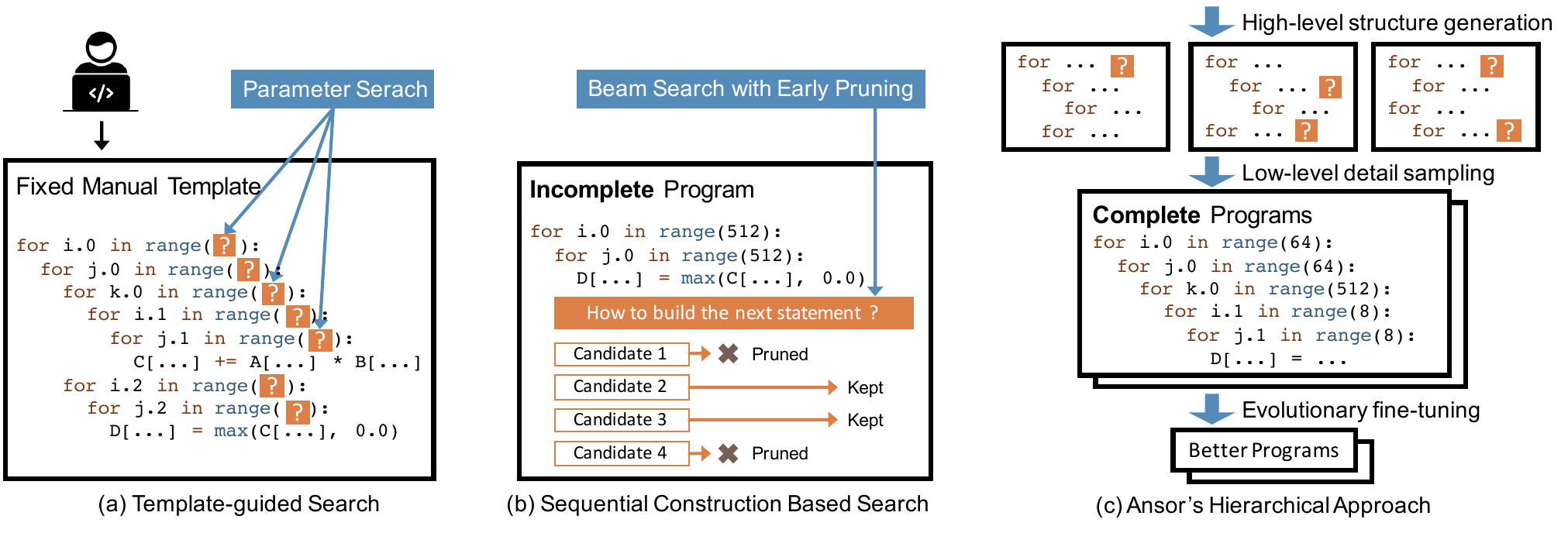}
	\precap
	\caption{Search strategy comparison. The pseudo-code shows tensor programs with loop nests. The question marks in orange background denote low-level parameters.}
	\postcap
	\label{fig:search-algorithm-comparison}
\end{figure*}

\textbf{Template-guided search.}
In template-guided search, the search space is defined by manual templates.
As shown in \autorefsuffix{fig:search-algorithm-comparison}{a}, the compiler (\textit{e.g.}, TVM) requires the user to manually write a template for a computation definition.
The template defines the structure of the tensor programs with some tunable parameters (\textit{e.g.}, tile size and unrolling factor).
The compiler then searches for the best values of these parameters for a specific input shape configuration and a specific hardware target. 
This approach has achieved good performance on common deep learning operators.
However, developing templates requires substantial effort.
For example, the code repository of TVM already contains more than 15K lines of code for these templates. This number continues to grow as new operators and new hardware platforms emerge.
Besides, constructing a quality template requires expertise in both tensor operators and hardware. It takes non-trivial research effort ~\cite{liu2019optimizing, wang2019unified, zheng2020flextensor} to develop quality templates.
Despite the complexity of template design, manual templates only cover limited program structures because manually enumerating all optimization choices for all operators is prohibitive.
This approach typically requires defining one template for each operator.
FlexTensor \cite{zheng2020flextensor} proposes a general template to cover multiple operators, but its template is still designed for single operator granularity, which fails to include optimizations involving multiple operators (\textit{e.g.}, operator fusion).
The search space of optimizing a computational graph with multiple operators should contain different ways to compose the operators. A template-based approach fails to achieve this because it cannot break down their fixed templates and re-compose them during the search.

\textbf{Sequential construction based search.} This approach defines the search space by decomposing the program construction into a fixed sequence of decisions.  The compiler then uses an algorithm such as beam search \cite{medress1977speech} to search for good decisions (\textit{e.g.}, Halide auto-scheduler~\cite{adams2019learning}). 
In this approach, the compiler constructs a tensor program by sequentially unfolding all nodes in the computational graph. 
For each node, the compiler makes a few decisions on how to transform it into low-level tensor programs (\textit{i.e.,} deciding computation location, storage location, tile size, \textit{etc.}).
When all nodes are unfolded, a complete tensor program is constructed.
This approach uses a set of general unfolding rules for every node, so it can search automatically without requiring manual templates.
Because the number of possible choices of each decision is large, to make the sequential process feasible, this approach keeps only top-$k$ candidate programs after every decision.
The compiler estimates and compares the performance of candidate programs with a learned cost model to select the top-$k$ candidates; while other candidates are pruned. 
During the search, the candidate programs are incomplete because only part of the computational graph is unfolded or only some of the decisions are made.
\autorefsuffix{fig:search-algorithm-comparison}{b} shows this process.

\begin{figure}[t]
	\centering
	\includegraphics[width=\columnwidth]{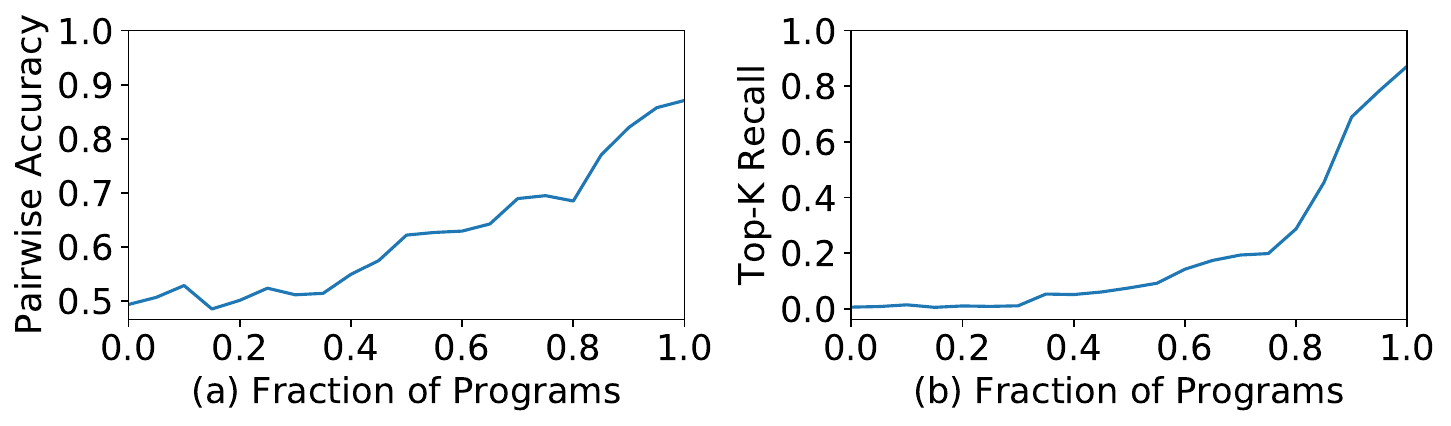}
	\precap
	\caption{Pairwise comparison accuracy and top-$k$ recall curve on random partial programs. In both subfigures, higher values are better.}
	\postcap
	\label{fig:partial-evaluation}
\end{figure}

However, estimating the final performance of incomplete programs is difficult in several respects:
(1) the cost model trained on complete programs cannot accurately predict the final performance of incomplete programs. The cost model can only be trained on complete programs because we need to compile programs and measure their execution time to get the labels for training.
Directly using this model to compare the final performance of incomplete programs will result in poor accuracy.
As a case study, we train our cost model (\autoref{subsec:learned-cost-model}) on 20,000 random complete programs from our search space and use the model to predict the final performance of incomplete programs. 
The incomplete programs are obtained by only applying a fraction of loop transformations of the complete programs.
We use two ranking metrics for evaluation: the accuracy of pairwise comparison and the recall@$k$ score of top-$k$ programs 
\footnote{{recall@$k$ of top-$k$ = $\frac{|G \cap P|}{k}$, where $G$ is the set of top-$k$ programs according to the ground truth and $P$ is the set of top-$k$ programs predicted by the model. \label{fn:top-k-recall-definition} }} ($k=10$).
As shown in \autoref{fig:partial-evaluation}, the two curves start from $50\%$ and $0\%$ respectively, meaning that random guess with zero information gives $50\%$ pairwise comparison accuracy and $0\%$ top-$k$ recall.
The two curves increase quickly as the programs become complete, which means the cost model performs very well for complete programs but fails to accurately predict the final performance of incomplete programs.
(2) The fixed order of sequential decisions limits the design of the search space.
For example, some optimization needs to add new nodes to the computational graph (\textit{e.g.}, adding cache nodes, using \texttt{rfactor}\cite{suriana2017parallel}). The number of decisions for different programs becomes different. It is hard to align the incomplete programs for a fair comparison.
(3) Sequential construction based search is not scalable. Enlarging the search space needs to add more sequential construction steps, which, however, leads to a worse accumulated error.

\textbf{Ansor's hierarchical approach} 
As shown in \autorefsuffix{fig:search-algorithm-comparison}{c}, \Ansor is backed by a hierarchical search space that decouples high-level structures and low-level details.
\Ansor constructs the search space for a computational graph automatically, eliminating the need to manually develop templates. \Ansor then samples complete programs from the space and performs fine-tuning on complete programs, avoiding the inaccurate estimation of incomplete programs.
\autoref{fig:search-algorithm-comparison} shows the key difference between \Ansor's approach and existing approaches.

\section{Design Overview}
\label{sec:overview}

\begin{figure}[t]
	\centering
	\includegraphics[width=0.9\columnwidth]{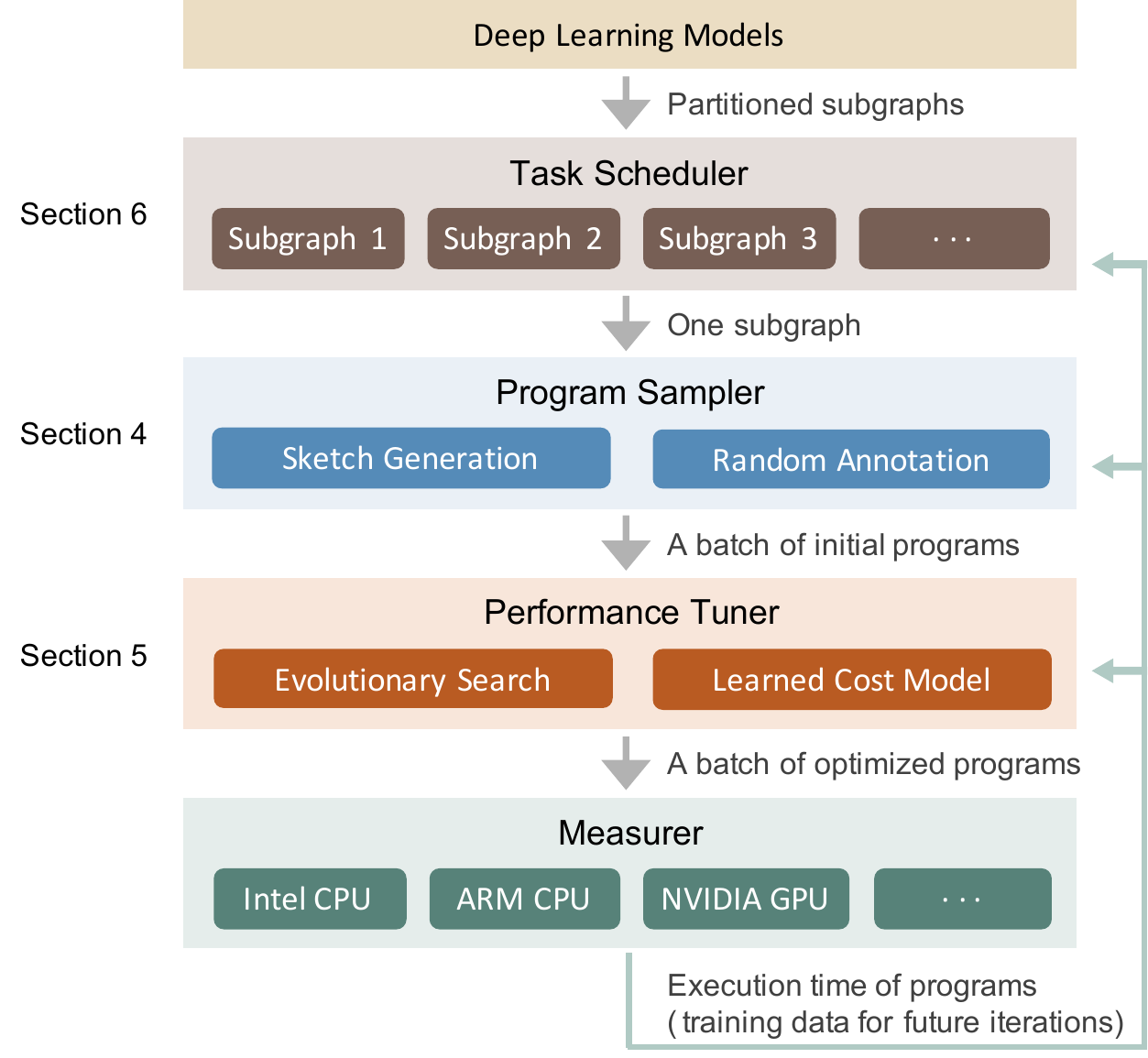}
	\vskip -0.8em
	\caption{System Overview.
		The gray arrows show the flow of extracting subgraphs from deep learning models and generating optimized programs for them.
		The green arrows mean the measurer returns profiling data to update the status of all components in the system.}
	\postcap 
	\label{fig:system-overview}
\end{figure}

\Ansor is an automated tensor program generation framework.
\autoref{fig:system-overview} shows the overall architecture of \Ansor. The input of \Ansor is a set of to be optimized DNNs.
Ansor uses the operator fusion algorithm from Relay\cite{roesch2019relay} to convert DNNs from popular model formats (\textit{e.g.}, ONNX~\cite{bai2019onnx}, TensorFlow PB) to partitioned small subgraphs.
\Ansor then generates tensor programs for these subgraphs.
\Ansor has three major components: (1) a program sampler that constructs a large search space and samples diverse programs from it; (2) a performance tuner that fine-tunes the performance of sampled programs; (3) a task scheduler that allocates time resources for optimizing multiple subgraphs in the DNNs.

\textbf{Program sampler.} One key challenge \Ansor has to address is generating a large search space for a given computational graph.
To cover diverse tensor programs with various high-level structures and low-level details, \Ansor utilizes a hierarchical representation of the search space with two levels: \textit{sketch} and \textit{annotation}   (\autoref{sec:program-sampling}). 
\Ansor defines the high-level structures of programs as sketches and leaves billions of low-level choices (\textit{e.g.}, tile size, parallel, unroll annotations) as annotations.
This representation allows \Ansor to enumerate high-level structures flexibly and sample low-level details efficiently.
\Ansor includes a program sampler that randomly samples programs from the space to provide comprehensive coverage of the search space. 

\textbf{Performance tuner.} The performance of randomly sampled programs is not necessarily good. The next challenge is to fine-tune them.
\Ansor employs evolutionary search and a learned cost model to perform fine-tuning iteratively (\autoref{sec:performance-fine-tuning}).
At each iteration, \Ansor uses re-sampled new programs as well as good programs from previous iterations as the initial population to start the evolutionary search.
Evolutionary search fine-tunes programs by mutation and crossover which perform out-of-order rewrite and address the limitation of sequential construction.
Querying the learned cost model is orders of magnitude faster than actual measurement, so we can evaluate thousands of programs in seconds.

\textbf{Task scheduler.} Using program sampling and performance fine-tuning allows \Ansor to find high-performance tensor programs for a computational graph.
Intuitively, treating a whole DNN as a single computational graph and generating a full tensor program for it could potentially achieve the optimal performance. This, however, is inefficient because it has to deal with the unnecessary exponential explosion of the search space.
Typically, the compiler partitions the large computational graph of a DNN into several small subgraphs~\cite{roesch2019relay, chen2018tvm}. This partition has a negligible effect on the performance thanks to the layer-by-layer construction nature of DNNs.
This brings the final challenge of \Ansor: how to allocate time resources when generating programs for multiple subgraphs.
The task scheduler  (\autoref{sec:task-scheduler}) in \Ansor uses a scheduling algorithm based on gradient descent to allocate resources to the subgraphs that are more likely to improve the end-to-end DNN performance.

\section{Program Sampling}
\label{sec:program-sampling}
The search space an algorithm explores determines the best programs it can find.
The considered search spaces in existing approaches are limited by the following factors: (1) Manual enumeration (\textit{e.g.}, TVM \cite{chen2018learning}). It is impractical to manually enumerate all possible choices by templates, so existing manual templates only cover a limited search space heuristically. (2) Aggressive early pruning (\textit{e.g.}, Halide auto-scheduler \cite{adams2019learning}). Aggressive early pruning based on evaluating incomplete programs prevents the search algorithm from exploring certain regions in the space.

In this section, we introduce techniques to push the boundary of the considered search space by addressing the above limitations. To solve (1), we automatically expand the search space by recursively applying a set of flexible derivation rules. To avoid (2), we randomly sample complete programs in the search space.
Since random sampling gives an equal chance to every point to be sampled, our search algorithm can potentially explore every program in the considered space.
We do not rely on random sampling to find the optimal program, because every sampled program is later fined-tuned (\autoref{sec:performance-fine-tuning}).

\begin{table*}[t!]
	\renewcommand{\arraystretch}{1.2}
	\footnotesize
	\centering
	\begin{tabular}{llll}
		\hline
		No & Rule Name & Condition & Application  \\
		\hline
		1 & Skip &  $\neg  IsStrictInlinable(S, i)$  & $S' = S;  i' = i - 1$ \\
		2 & Always Inline  & $IsStrictInlinable(S, i) $  & $S' = Inline(S, i)$; $i' = i - 1$ \\
		3 & Multi-level Tiling & $HasDataReuse(S, i)$ & $S' = MultiLevelTiling(S, i); i' = i - 1$ \\
		4 & Multi-level Tiling with Fusion& $HasDataReuse(S, i) \wedge HasFusibleConsumer(S, i) $  & $S' = FuseConsumer(MultiLevelTiling(S, i),i); i' = i - 1$ \\
		5 & Add Cache Stage & $HasDataReuse(S, i) \wedge \neg HasFusibleConsumer(S, i) $  & $S' = AddCacheWrite(S, i); i = i' $ \\
		6 & Reduction Factorization &  $HasMoreReductionParallel(S, i)$ & $S' = AddRfactor(S, i); i' = i - 1 $  \\
		... & User Defined Rule &  ... & ...  \\
		\hline
	\end{tabular}
	\vskip -0.4em
	\caption{Derivation rules used to generate sketches. The condition runs on the current state $\sigma = (S, i)$. The application derives the next state $\sigma' = (S', i')$ from the current state $\sigma$. Note that some function (\textit{e.g.}, $AddRfactor$, $FuseConsumer$) can return multiple possible values of $S'$. In this case we collect all possible $S'$, and return multiple next states $\sigma'$ for a single input state $\sigma$.}
	\postcap
	\label{table:derivation-rules}
\end{table*}

To sample programs that can cover a large search space, we define a hierarchical search space with two levels: \textit{sketch} and \textit{annotation}.
We define the high-level structures of programs as sketches and leave billions of low-level choices (\textit{e.g.}, tile size, parallel, unroll annotations) as annotations.
At the top level, we generate sketches by recursively applying a few derivation rules.
At the bottom level, we randomly annotate these sketches to get complete programs.
This representation summarizes a few basic structures from billions of low-level choices, enabling the flexible enumeration of high-level structures and efficient sampling of low-level details.

While \Ansor supports both CPU and GPU, we explain the sampling process for CPUs in \autoref{subsec:sketch-generation} and \autoref{subsec:random-sampling} as an example.
We then discuss how the process is different for GPU in \autoref{subsec:gpu-support}.

\subsection{Sketch Generation}
\label{subsec:sketch-generation}
\begin{figure*}[t!]
	\centering
	\includegraphics[width=\textwidth]{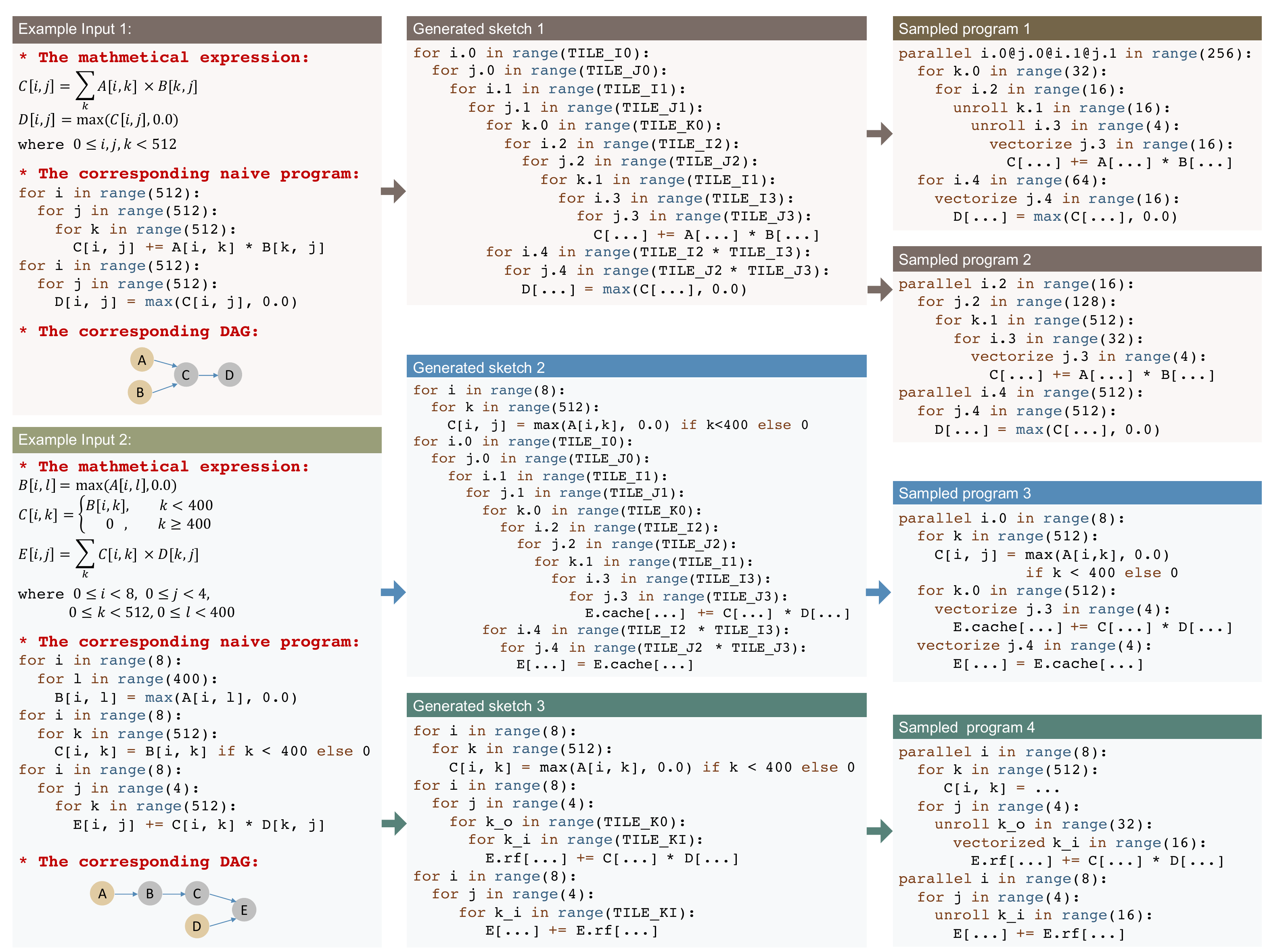}
	\vskip -0.8em
	\caption{Examples of generated sketches and sampled programs. This figure shows two example inputs, three generated sketches and four sampled programs. The code example is pseudo code in a python-like syntax.}
	\postcap
	\label{fig:sampled-program-examples}
\end{figure*}

As shown in \autoref{fig:system-overview}, the program sampler accepts partitioned subgraphs as input.
The first column in \autoref{fig:sampled-program-examples} shows two examples of the input.
The input has three equivalent forms: the mathematical expression, the corresponding naive program obtained by directly expanding the loop indices, and the corresponding computational graph (directed acyclic graph, or DAG).

To generate sketches for a DAG with multiple nodes, we visit all the nodes in a topological order and build the structure iteratively.
For computation nodes that are compute-intensive and have a lot of data reuse opportunities (\textit{e.g.}, conv2d, matmul), we build basic tile and fusion structures for them as the sketch.
For simple element-wise nodes (\textit{e.g.}, ReLU, element-wise add), we can safely inline them.
Note that new nodes (\textit{e.g.}, caching nodes, layout transform nodes) may also be introduced to the DAG during the sketch generation.

We propose a derivation-based enumeration approach to generate all possible sketches by recursively applying several basic rules.
This process takes a DAG as an input and returns a list of sketches.
We define the State $\sigma = (S, i)$, where $S$ is the current partially generated sketch for the DAG, and $i$ is the index of the current working node.
The nodes in a DAG are sorted in a topological order from output to input.
The derivation begins from the initial naive program and the last node, or the initial state $\sigma = (naive\ program, index\ of \ the\ last\ node)$.
Then we try to apply all derivation rules to the states recursively.
For each rule, if the current state satisfies the application condition, we apply the rule to $\sigma = (S, i)$ and get $\sigma' = (S', i')$ where $i' \le i$.
This way the index $i$ (working node) decreases monotonically.
A state becomes a terminal state when $i = 0$.
During enumeration, multiple rules can be applied to one state to generate multiple succeeding states. One rule can also generate multiple possible succeeding states. So we maintain a queue to store all intermediate states. The process ends when the queue is empty.
All $\sigma.S$ in terminal states form a sketch list at the end of the sketch generation. The number of sketches is less than 10 for a typical subgraph.

\textbf{Derivation rules.}
\autoref{table:derivation-rules} lists derivation rules we used for the CPU.
We first provide the definition of the used predicates and then describe the functionality of each rule.
$IsStrictInliable(S, i)$ indicates if the node $i$ in $S$ is a simple element-wise operator that can always be inlined (\textit{e.g.}, element-wise add, ReLU). $HasDataReuse(S, i)$ indicates if the node $i$ in $S$ is a compute-intensive operator and has plentiful intra-operator data reuse opportunity (\textit{e.g.}, matmul, conv2d). $HasFusibleConsumer(S, i)$ indicates if the node $i$ in $S$ has only one consumer $j$ and node $j$ can be fused into node $i$ (\textit{e.g.}, matmul + bias\_add, conv2d + relu). $HasMoreReductionParallel(S, i)$ indicates if the node $i$ in $S$ has little parallelism in space dimensions but has ample parallelism opportunity in reduction dimensions.  (\textit{e.g.}, computing 2-norm of a matrix, matmul $C_{2\times 2} = A_{2\times 512} \cdot B_{512 \times 2}$).
We perform static analysis on the computation definitions to get the values for these predicates.
The analysis is done automatically by parsing the read/write pattern in the mathematical expressions.
Next, we introduce the functionality of each derivation rule.

Rule 1 just simply skips a node if it is not strictly inlinable. Rule 2 always inlines strictly inlinable nodes. Since the conditions of rule 1 and rule 2 are mutually exclusive, a state with $i>1$ can always satisfy one of them and continue to derive.

Rules 3, 4, and 5 deal with the multi-level tiling and fusion for nodes that have data reuse.
Rule 3 performs multi-level tiling for data reusable nodes. For CPU, we use a ``SSRSRS'' tile structure, where ``S'' stands for one tile level of space loops and ``R'' stands for one tile level of reduction loops. For example, in the matmul $C(i, j) = \sum_k A[i,k] \times B[k, j]$, $i$ and $j$ are space loops and $k$ is a reduction loop. The ``SSRSRS'' tile structure for matmul expands the original 3-level loop $(i, j, k)$ into a 10-level loop $(i_0, j_0, i_1, j_1, k_0, i_2, j_2, k_1, i_3, j_3)$. Although we do not permute the loop order, this multi-level tiling can also cover some cases of reordering.
For example, the above 10-level loop can be specialized to just a simple reorder $(k_0, j_2, i_3)$ by setting the length of other loops to one.
The "SSRSRS" tile structure is general for compute-intensive dense operators (\textit{e.g.}, matmul, conv2d, conv3d) in deep learning, because they all consist of space loops and reduction loops.

Rule 4 performs multi-level tiling and also fuses the fusible consumers. For example, we fuse the element-wise nodes (\textit{e.g.}, ReLU, bias add) into the tiled nodes (\textit{e.g.}, conv2d, matmul).
Rule 5 adds a caching node if the current data-reusable node does not have a fusible consumer.
For example, the final output node in a DAG does not have any consumer, so it directly writes results into main memory by default and this is inefficient due to the high latency of memory accesses.
By adding a cache node, we introduce a new fusible consumer into the DAG, then rule 4 can be applied to fuse this newly added cache node into the final output node.
With the cache node fused, now the final output node writes its results into a cache block, and the cache block will be written to the main memory at once when all data in the block is computed.

Rule 6 can use \texttt{rfactor} \cite{suriana2017parallel} to factorize a reduction loop into a space loop to bring more parallelism.

\textbf{Examples.}
\autoref{fig:sampled-program-examples} shows three examples of the generated sketches.
The sketches are different from the manual templates in TVM, because the manual templates specify both high-level structures and low-level details while sketches only define high-level structures.
For the example input 1, the sorted order of the four nodes in the DAG is $(A, B, C, D)$.
To derive the sketches for the DAG, we start from output node $D (i=4)$ and apply rules to the nodes one by one.
Specifically, the derivation for generated sketch 1 is:
\preeq \preeq \preeq
\begin{align*}
Input~1  \rightarrow  & \sigma(S_0, i = 4)  \xrightarrow{\text{Rule 1}}  \sigma(S_1, i = 3)  \xrightarrow{\text{Rule 4}}  \\ 
 & \sigma(S_2, i = 2)   \xrightarrow{\text{Rule 1}}  \sigma(S_3, i = 1)  \xrightarrow{\text{Rule 1}}  Sketch~1
\end{align*}
\posteq

For the example input 2, the sorted order of the five nodes is $(A, B, C, D, E)$.
Similarly, we start from the output node $E (i=5)$ and apply rules recursively.
The generated sketch 2 is derived by:
\preeq \preeq \preeq
\begin{align*}
 Input~2 \rightarrow  & \sigma(S_0, i = 5)  \xrightarrow{\text{Rule 5}}  \sigma(S_1, i = 5)  \xrightarrow{\text{Rule 4}} \\
 & \sigma(S_2, i = 4)  \xrightarrow{\text{Rule 1}} \sigma(S_3, i = 3) \xrightarrow{\text{Rule 1}} \\
 & \sigma(S_4, i = 2)  \xrightarrow{\text{Rule 2}} \sigma(S_5, i = 1)  \xrightarrow{\text{Rule 1}}  Sketch~2
\end{align*}
\posteq

\noindent Similarly, the generated sketch 3 is derived by:

\preeq \preeq
\begin{align*}
Input~2 \rightarrow  & \sigma(S_0, i = 5)  \xrightarrow{\text{Rule 6}}  \sigma(S_1, i = 4)  \xrightarrow{\text{Rule 1}} \\
& \sigma(S_2, i = 3)  \xrightarrow{\text{Rule 1}} \sigma(S_3, i = 2) \xrightarrow{\text{Rule 2}} \\
& \sigma(S_4, i = 1)  \xrightarrow{\text{Rule 1}}  Sketch~3
\end{align*}
\posteq

\textbf{Customization.}
While the presented rules are practical enough to cover the structures for most operators, there are always exceptions.
For example, some special algorithms (\textit{e.g.}, Winograd convolution~\cite{lavin2016fast}) and accelerator intrinsics (\textit{e.g.}, TensorCore~\cite{nvidia2017tensorcore}) require special tile structures to be effective.
Although the template-guided search approach (in TVM) can craft a new template for every new case, it needs a great amount of design effort.
On the other hand, the derivation-based sketch generation in \Ansor is flexible enough to generate the required structures for emerging algorithms and hardware, as we allow users to register new derivation rules and integrate them seamlessly with existing rules.

\subsection{Random Annotation}
\label{subsec:random-sampling}

The sketches generated by the previous subsection are incomplete programs because they only have tile structures without specific tile sizes and loop annotations, such as parallel, unroll, and vectorization.
In this subsection, we annotate sketches to make them complete programs for fine-tuning and evaluation.

Given a list of generated sketches, we randomly pick one sketch, randomly fill out tile sizes, parallelize some outer loops, vectorize some inner loops, and unroll a few inner loops.
We also randomly change the computation location of some nodes in the program to make a slight tweak to the tile structure.
All ``random'' in this subsection means a uniform distribution over all valid values.
If some special algorithms require custom annotations to be effective (\textit{e.g.}, special unrolling), we allow users to give simple hints in the computation definition to adjust the annotation policy.
Finally, since changing the layout of constant tensors can be done in compilation time and brings no runtime overhead, we rewrite the layouts of the constant tensors according to the multi-level tile structure to make them as cache-friendly as possible.
This optimization is effective because the weight tensors of convolution or dense layers are constants for inference applications.

Examples of random sampling are shown in \autoref{fig:sampled-program-examples}.
The sampled program might have fewer loops than the sketch because the loops with length one are simplified.

\subsection{GPU Support}
\label{subsec:gpu-support}

For GPU, we change the  multi-level tiling structure from "SSRSRS" to "SSSRRSRS" to match the architecture of GPU. The loops in the first three space tiles are bound to BlockIdx, virtual thread (for reducing bank conflicts), and ThreadIdx, respectively. We add two sketch derivation rules, one for utilizing shared memory by inserting a caching node (similar to Rule 5) and the other for cross-thread reduction (similar to Rule 6).

\section{Performance Fine-tuning}
\label{sec:performance-fine-tuning}
\postsec

The programs sampled by the program sampler have good coverage of the search space, but their qualities are not guaranteed.
This is because the optimization choices, such as tile structure and loop annotations, are all randomly sampled.
In this section, we introduce the performance tuner that fine-tunes the performance of the sampled programs via evolutionary search and a learned cost model.

The fine-tuning is performed iteratively. At each iteration, we first use evolutionary search to find a small batch of promising programs according to a learned cost model.  We then measure these programs on hardware to get the actual execution time cost. Finally, the profiling data got from measurement is used to re-train the cost model to make it more accurate.

The evolutionary search uses randomly sampled programs as well as high-quality programs from the previous measurement as the initial population and applies mutation and crossover to generate the next generation. The learned cost model is used to predict the \textit{fitness} of each program, which is the throughput of one program in our case.
We run evolution for a fixed number of generations and pick the best programs found during the search.
We leverage a learned cost model because the cost model can give relatively accurate estimations of the fitness of programs while being orders of magnitudes faster than the actual measurement.
It allows us to compare tens of thousands of programs in the search space in seconds, and pick the promising ones to do actual measurements.

\subsection{Evolutionary Search}
\label{subsec:evolutionary-search}

Evolutionary search~\cite{vikhar2016evolutionary} is a generic meta-heuristic algorithm inspired by biological evolution. By iteratively mutating high-quality programs, we can generate new programs with potentially higher quality. The evolution starts from the sampled initial generation.
To generate the next generation, we first select some programs from the current generation according to certain probabilities. The probability of selecting a program is proportional to its fitness predicted by the learned cost model (\autoref{subsec:learned-cost-model}), meaning that the program with a higher performance score has a higher probability to be selected.
For the selected programs, we randomly apply one of the evolution operations to generate a new program. Basically, for decisions we made during sampling (\autoref{subsec:random-sampling}), we design corresponding evolution operations to rewrite and fine-tune them.

\textbf{Tile size mutation.}
This operation scans the program and randomly selects a tiled loop.
For this tiled loop, it divides a tile size of one tile level by a random factor and multiplies this factor to another level.
Since this operation keeps the product of tile sizes equal to the original loop length, the mutated program is always valid.

\textbf{Parallel mutation.}
This operation scans the program and randomly selects a loop that has been annotated with parallel. 
For this loop, this operation changes the parallel granularity by either fusing its adjacent loop levels or splitting it by a factor.

\textbf{Pragma mutation.}
Some optimizations in a program are specified by compiler-specific pragma. This operation scans the program and randomly selects a pragma. For this pragma, this operation randomly mutates it into another valid value. For example, our underlying code generator supports auto unrolling with a maximum number of steps by providing an \texttt{auto\_unroll\_max\_step=N} pragma. We randomly tweak the number $N$.

\textbf{Computation location mutation.}
This operation scans the program and randomly selects a flexible node that is not multi-level tiled (\textit{e.g.}, a padding node in the convolution layer). For this node, the operation randomly changes its computation location to another valid attach point.

\textbf{Node-based crossover.}
Crossover is an operation to generate new offspring by combining the genes from two or more parents.
The genes of a program in \Ansor are its rewriting steps.
Every program generated by \Ansor is rewritten from its initial naive implementation.
\Ansor preserves a complete rewriting history for each program during sketch generation and random annotation.
We can treat rewriting steps as the genes of a program because they describe how this program is formed from the initial naive one.
Based on this, we can generate a new program by combining the rewriting steps of two existing programs.
However, arbitrarily combining rewriting steps from two programs might break the dependencies in steps and create an invalid program.
As a result, the granularity of crossover operation in \Ansor is based on nodes in the DAG, because the rewriting steps across different nodes usually have less dependency.
Ansor randomly selects one parent for each node and merges the rewriting steps of selected nodes.
When there are dependencies between nodes, \Ansor tries to analyze and adjust the steps with simple heuristics. \Ansor further verifies the merged programs to guarantee the functional correctness. The verification is simple because \Ansor only uses a small set of loop transformation rewriting steps, and the underlying code generator can check the correctness by dependency analysis.

The evolutionary search leverages mutation and crossover to generate a new set of candidates repeatedly for several rounds and outputs a small set of programs with the highest scores.
These programs will be compiled and measured on the target hardware to obtain the real running time cost.
The collected measurement data is then used to update the cost model.
In this way, the accuracy of the learned cost model is gradually improved to match the target hardware.
Consequently, the evolutionary search gradually generates higher-quality programs for the target hardware platform.

Unlike the search algorithms in TVM and FlexTensor that can only work in a fixed grid-like parameter space, the evolutionary operations in \Ansor are specifically designed for tensor programs. They can be applied to general tensor programs and can handle a search space with complicated dependency.
Unlike the unfolding rules in Halide auto-scheduler, these operations can perform out-of-order modifications to programs, addressing the sequential limitations.

\subsection{Learned Cost Model}
\label{subsec:learned-cost-model}

A cost model is necessary for estimating the performance of programs quickly during the search. We adopt a learned cost model similar to related works \cite{adams2019learning, chen2018learning} with newly designed program features.
A system based on learned cost models has great portability because a single model design can be reused for different hardware backends by feeding in different training data.

Since our target programs are mainly data parallel tensor programs, which are made by multiple interleaved loop nests with several assignment statements as the innermost statements, we train the cost model to predict the score of one innermost non-loop statement in a loop nest.
For a full program, we make predictions for each innermost non-loop statement and add the predictions up as the score. We build the feature vector for an innermost non-loop statement by extracting features in the context of a full program. The extracted features include arithmetic features and memory access features.
A detailed list of extracted features is in \autoref{sec:extracted-features}.

We use weighted squared error as the loss function. Because we mostly care about identifying the well-performing programs from the search space, we put more weight on the programs that run faster.
Specifically, the loss function of the model $f$ on a program $P$ with throughput $y$ is 
$ loss(f, P, y) = w_p (\sum_{s \in S(P)} f(s) - y )^2 =  y (\sum_{s \in S(P)} f(s) - y)^2  $
where $S(P)$ is the set of innermost non-loop statements in $P$. We directly use the throughput $y$ as weight.
We train a gradient boosting decision tree \cite{chen2016xgboost} as the underlying model $f$. A single model is trained for all tensor programs coming from all DAGs, and we normalize the throughput of all programs coming from the same DAG to be in the range of $[0, 1]$.
When optimizing a DNN, the number of measured programs are typically less than 30,000.
Training a gradient boosting decision tree is very fast on such a small data sets, so we train a new model every time instead of doing incremental updates.

\section{Task Scheduler}
\label{sec:task-scheduler}

A DNN can be partitioned into many independent subgraphs (\textit{e.g.}, conv2d + relu). For some subgraphs, spending time in tuning them does not improve the end-to-end DNN performance significantly. This is due to two reasons: either (1) the subgraph is not a performance bottleneck, or (2) tuning brings only minimal improvement in the subgraph's performance.

To avoid wasting time on tuning unimportant subgraphs, \Ansor dynamically allocates different amounts of time resources to different subgraphs.
Take ResNet-50 for example, it has 29 unique subgraphs after the graph partitioning.
Most of these subgraphs are convolution layers with different shapes configurations (input size, kernel size, stride, etc). We need to generate different programs for different convolution layers because the best tensor program depends on these shape configurations. 
In reality, users may have multiple DNNs for all their applications. This leads to more subgraphs as well as more opportunities to reduce the total tuning time, because we can share and reuse knowledge between subgraphs.
A subgraph can also appear multiple times in a DNN or across different DNNs.

We define a task as a process performed to generate high-performance programs for a subgraph.
It means that optimizing a single DNN requires finishing dozens of tasks (\textit{e.g.}, 29 tasks for ResNet-50).
\Ansor's task scheduler allocates time resources to tasks in an iterative manner.
At each iteration, \Ansor selects a task, generates a batch of promising programs for the subgraph, and measures the program on hardware.
We define such an iteration as one unit of time resources. When we allocate one unit of time resources to a task, the task obtains an opportunity to generate and measure new programs, which also means the chance to find better programs.
We next present the formulation of the scheduling problem and our solution.

\subsection{Problem Formulation}
\label{subsec:task-scheduler-problem}
When tuning a DNN or a set of DNNs, a user can have various types of goals, for example, reducing a DNN's latency, meeting latency requirements for a set of DNNs, or minimizing tuning time when tuning no longer improves DNN performance significantly. We thus provide users a set of objective functions to express their goals. Users can also provide their own objective functions. 

Suppose there are $n$ tasks in total. Let $t \in \mathbf{Z}^n$ be the allocation vector, where $t_i$ is the number of time units spent on task $i$. Let the minimum subgraph latency task $i$ achieves be a function of the allocation vector $g_i(t)$. Let the end-to-end cost of the DNNs be a function of the latency of the subgraphs $f(g_1(t), g_2(t), ..., g_3(t))$. Our objective is to minimize the end-to-end cost:
\preeq \preeq
$$ \text{minimize } f(g_1(t), g_2(t), ..., g_3(t)) $$
\posteq

To minimize the end-to-end latency of a single DNN, we can define $f(g_1, g_2, ..., g_n) = \sum_{i=1}^{n} {w_i \times g_i}$,
where $w_i$ is the number of appearances of task $i$ in the DNN. This formulation is straightforward because $f$ is an approximation of the end-to-end DNN latency.

When tuning a set of DNNs, there are several options.
\autoref{table:objective-function-multiple-network} shows a number of example objective functions for tuning multiple DNNs.
Let $m$ be the number of DNNs, $S(j)$ is the set of tasks that belong to DNN $j$.
$f_1$ adds up the latency of every DNN, which means to optimize the cost of a pipeline that sequentially runs all DNNs once.
In $f_2$, we define $L_j$ as the latency requirement of DNN $j$, meaning that we do not want to spend time on a DNN if its latency has already met the requirement.
In $f_3$, we define $B_j$ as the reference latency of a DNN $j$. As a result, our goal is to maximize the geometric mean of speedup against the given reference latency.
Finally in $f_4$, we define a function $ES(g_i, t)$ that returns an early stopping value by looking at the history of latency of task $i$. It can achieve the effect of per-task early stopping.

\begin{table}[t!]
	\renewcommand{\arraystretch}{1.4}
	\centering
	\begin{tabular}{ll}
		\hline
		$f_1 =\sum_{j=1}^{m} \sum_{i \in S(j)} w_i \times g_i(t) $ \\ 
		$f_2 = \sum_{j=1}^{m} { \max( \sum_{i \in S(j)} w_i \times g_i(t), L_j)}$ \\
        $f_3 = -(\prod_{j=1}^{m} { \frac{B_j}{\sum_{i \in S(j)} w_i \times g_i(t)} }) ^ {\frac{1}{m}} $ \\
        $f_4 = \sum_{j=1}^{m} { \sum_{i \in S(j)} w_i \times \max( g_i(t), ES(g_i, t))}$ \\
		\hline
	\end{tabular}
	\caption{Examples of objective functions for multiple neural networks}
\postcap
	\label{table:objective-function-multiple-network}
\end{table}

\subsection{Optimizing with Gradient Descent}
\label{subsec:grdient-descent}
We propose a scheduling algorithm based on gradient descent to efficiently optimize the objective function.
Given the current allocation $t$, the idea is to approximate the gradient of the objective function $\frac{\partial f}{\partial t_i}$ in order to choose the task $i$ such that $i = \operatorname*{argmax}_i{|\frac{\partial f}{\partial t_i}|}$.
We approximate the gradient by making an optimistic guess and considering the similarity between tasks.

The derivation is in \autoref{sec:gradient-approximation}.
We approximate the gradient by 

\preeq \preeq \preeq
\begin{align*}
\frac{\partial f}{\partial t_i} 
&\approx \frac{\partial f}{\partial g_i} (\alpha \frac{g_i(t_i) - g_i(t_i - \Delta t)}{\Delta t}  
+ \\
& (1-\alpha)  (\min( - \frac{g_i(t_i)}{t_i}, \beta \frac{C_i}{\max_{k\in N(i)} {V_k}}  - g_i(t_i)))) \\
\end{align*}
\posteq \posteq \posteq \posteq

\noindent where $\Delta t$ is a small backward window size, $g_i(t_i)$ and $g_i(t_i - \Delta t)$ are known from the history of allocations. $N(i)$ is the set of similar tasks of $i$, $C_i$ is the number of floating point operation in task $i$ and $V_k$ is the number of floating point operation per second we can achieve in task $k$. 
The parameter $\alpha$ and  $\beta$ control the weight to trust some predictions.

To run the algorithm, \Ansor starts from $t = \mathbf{0}$ and warms up with a round of round-robin to get an initial allocation vector $t = (1, 1, ..., 1)$.
After the warm-up, at each iteration, we compute the gradient of each task and pick $\operatorname*{argmax}_i{|\frac{\partial f}{\partial t_i}|}$.
Then we allocate the resource unit to task $i$ and update the allocation vector $t_i = t_i + 1$.
The optimization process continues until we run out of the time budget.
To encourage exploration, we adopt a $\epsilon$-greedy strategy \cite{sutton2018reinforcement}, which preserves a probability of $\epsilon$ to randomly select a task.

Taking the case of optimizing for a single DNN's end-to-end latency for example, \Ansor prioritizes a subgraph that has a high initial latency because our optimistic guess says we can reduce its latency quickly. Later, if \Ansor spends many iterations on it without observing a decrease in its latency, \Ansor leaves the subgraph because its $|\frac{\partial f}{\partial t_i}|$ decreases.
\section{Evaluation}
\label{sec:evaluation}

The core of \Ansor is implemented in C++ with about 12K lines of code (3K for the search policy and 9K for other infrastructure).
\Ansor generates programs in its own intermediate representation (IR). 
These programs are then lowered to TVM IR for code generation targeting various hardware platforms. \Ansor only utilizes TVM as a deterministic code generator.

We evaluate the performance of generated programs on three levels: single operator, subgraph, and entire neural network.
For each level of evaluation, we compare \Ansor against the state-of-the-art search frameworks and hardware-specific manual libraries.
We also evaluate the search efficiency and the effectiveness of each component in \Ansor.

The generated tensor programs are benchmarked on three hardware platforms: an Intel CPU (18-core Platinum 8124M@3.0 GHz), an NVIDIA GPU (V100), and an ARM CPU (4-core Cortex-A53@1.4GHz on the Raspberry Pi 3b+). We use float32 as the data type for all evaluations.

\subsection{Single Operator Benchmark}
\label{subsec:single-op-benchmark}

\textbf{Workloads.}
We first evaluate \Ansor on a set of common deep learning operators, including 1D, 2D, and 3D convolution (C1D, C2D, and C3D respectively), matrix multiplication (GMM), group convolution (GRP), dilated convolution (DIL)~\cite{yu2015multi}, depth-wise convolution (DEP)~\cite{howard2017mobilenets}, transposed 2D convolution (T2D)~\cite{radford2015unsupervised}, capsule 2D convolution (CAP)~\cite{hinton2018matrix}, and matrix 2-norm (NRM). For each operator, we select 4 common shape configurations and evaluate them with two batch sizes (1 and 16).
In total, there are $10$ operators $\times 4$ shape configurations  $\times 2$  batch size $=80$ test cases.
The shape configurations used can be found in \autoref{sec:shape-eval}.
We run these test cases on the Intel CPU.

\textbf{Baselines.}
We include PyTorch (v1.5)\cite{paszke2019pytorch}, Halide auto-scheduler (commit: 1f875b0)\cite{adams2019learning}, FlexTensor (commit: 7ac302c)\cite{zheng2020flextensor}, and AutoTVM (commit: 69313a7)\cite{chen2018learning} as baselines.
PyTorch is backed by the vendor-provided kernel library MKL-DNN \cite{intel2017mkldnn}.
Halide auto-scheduler is a sequential construction based search framework for Halide. 
AutoTVM and FlexTensor are template-guided search frameworks based on TVM. 
Since Halide auto-scheduler does not have a pre-trained cost model for AVX-512, we disabled AVX-512 for the evaluation in \autoref{subsec:single-op-benchmark} and \autoref{subsec:subgraph-benchmark}.
For every operator, we use the best layout available in each framework, but the input and output tensors must not be packed.

\textbf{Search settings.}
We let search frameworks (\textit{i.e.}, Halide auto-scheduler, FlexTensor, AutoTVM, and \Ansor) to run search or auto-tuning with up to $1,000$ measurement trials per test case.
This means each framework can measure at most $80 \times 1,000$ programs for auto-tuning in this evaluation.
Using the same number of measurement trials makes it a fair comparison without involving implementation details.
In addition, using $1,000$ measurement trials per test case is typically enough for the search to converge in these frameworks.

\begin{figure}[t!]
	\centering
	\includegraphics[width=\columnwidth]{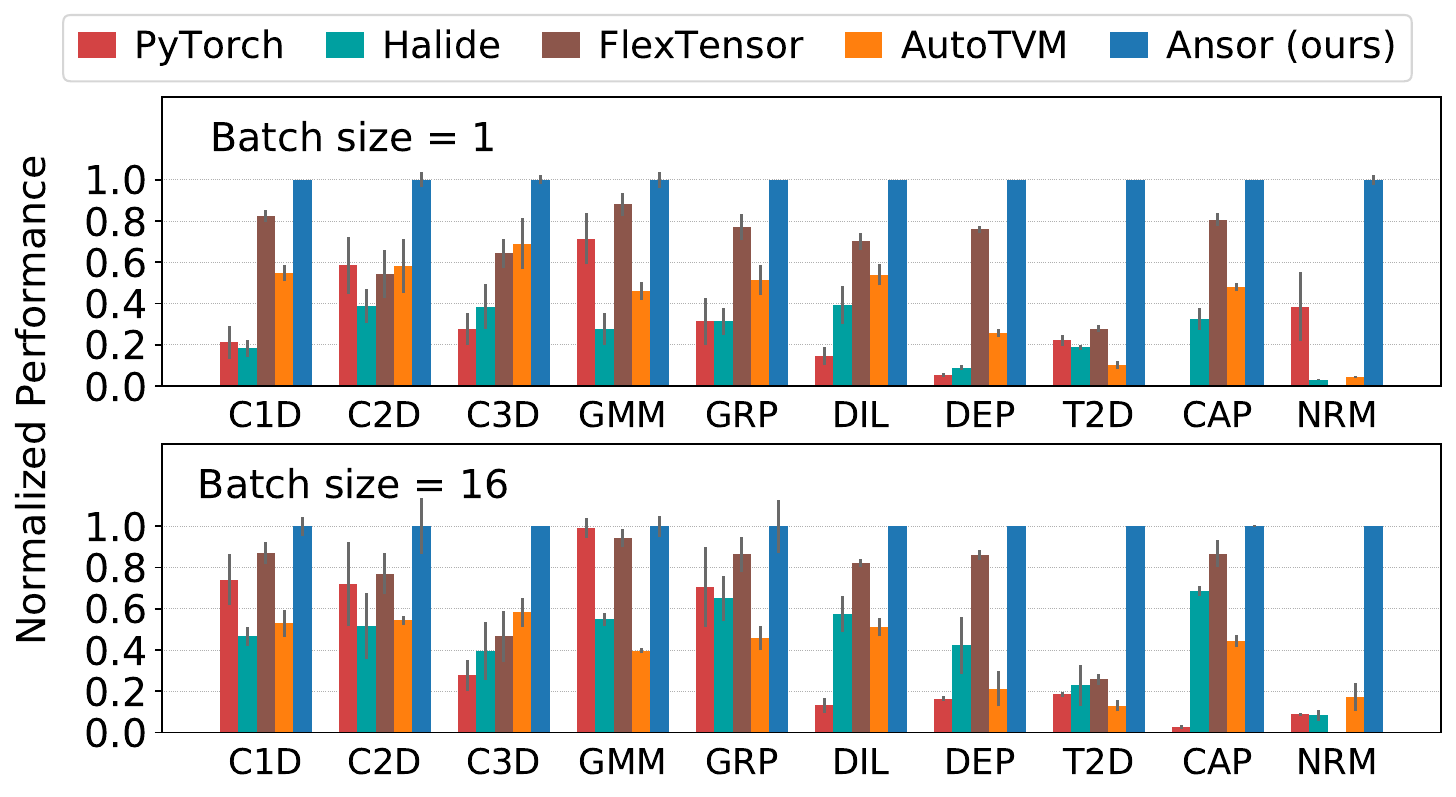}
	\precap
\caption{Single operator performance benchmark on a 20-core Intel-Platinum-8269CY. The y-axis is the throughput normalized to the best throughput for each operator.}
	\label{fig:cpu-single-op}
\end{figure}

\textbf{Normalization.}
\autoref{fig:cpu-single-op} shows the normalized performance.
For each test case, we normalize the throughputs to the best performing framework. We then plot the geometric mean of the four shapes of each operator. The geometric mean is also normalized to the best performing framework, so the best framework has a normalized performance of 1 in the figure. The error bar denotes the standard deviation of the normalized throughput of four shapes of each operator.

\textbf{Results.}
As shown in the \autoref{fig:cpu-single-op}, \Ansor performs the best or equally the best in all operator and batch size settings.
\Ansor outperforms existing search frameworks by  $1.1-22.5 \times$.
The performance improvements of \Ansor come from both its large search space and effective exploration strategy.
For most operators, we found the best program generated by \Ansor is outside the search space of existing search frameworks because \Ansor is able to explore more optimization combinations.
For example, the significant speedup on NRM is because \Ansor can parallelize reduction loops, while other frameworks do not.
The large speedup on T2D is because \Ansor can use correct tile structures and unrolling strategies to let the code generator simplify the multiplication of zeros in strided transposed convolution.
In contrast, other frameworks fail to capture many effective optimizations in their search space, making them unable to find the programs that \Ansor does.
For example, the unfolding rules in Halide do not split the reduction loop in GMM and do not split reduction loops in C2D when padding is computed outside of reduction loops.
The templates in AutoTVM have limited tile structures, as they cannot cover the structure of ``Generated Sketch 1'' in \autoref{fig:sampled-program-examples}.
The template in FlexTensor does not change the computation location of padding. The template in FlexTensor fails to run for reduction operators like NRM.

\begin{figure}[t!]
	\centering
	\includegraphics[width=\columnwidth]{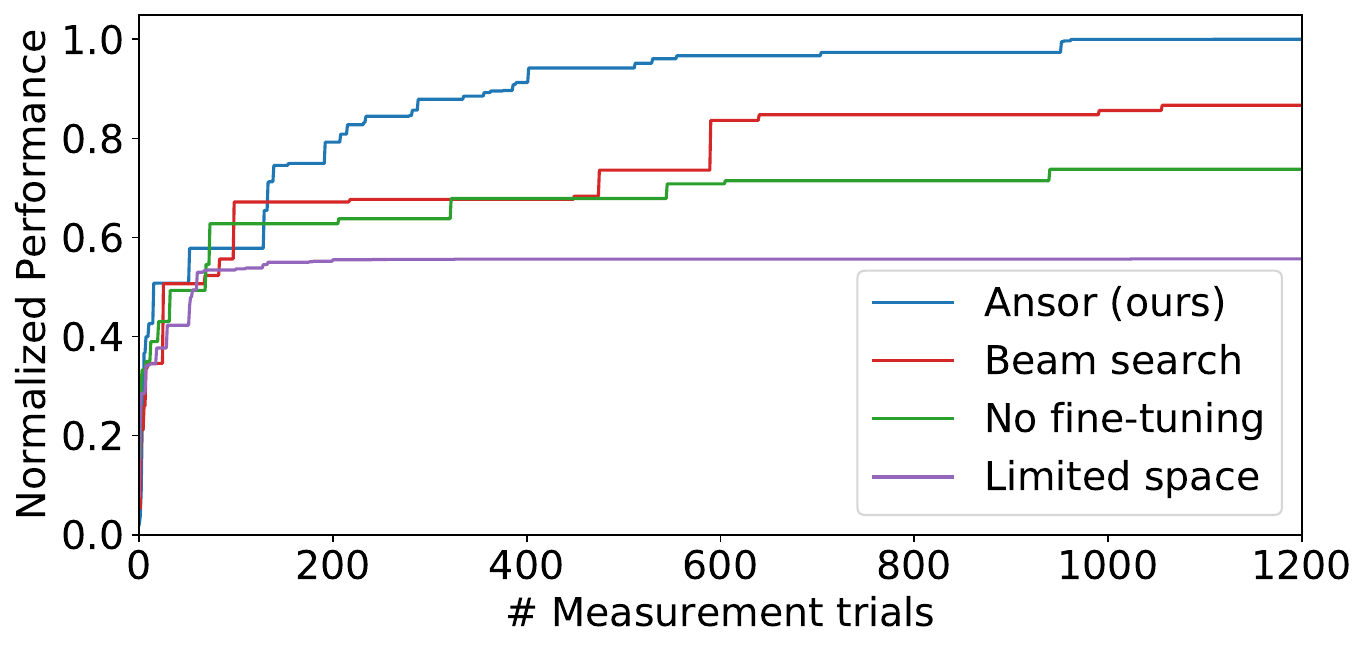}
	\precap
	\caption{Ablation study of four variants of \Ansor on a convolution operator. The y-axis is the throughput relative to the throughput of the best program.}
	\postcap 
	\label{fig:cpu-op-ablation}
\end{figure}

\textbf{Ablation study.} We run four variants of \Ansor on a convolution operator and report the performance curve.
We pick the last convolution operator in ResNet-50 with batch size=16 as the test case, because its search space is sufficiently large to evaluate the search algorithms. Other operators share a similar pattern.  
In \autoref{fig:cpu-op-ablation}, each curve is the median of 5 runs.
``\Ansor (ours)'' uses all our introduced techniques.
``Beam Search'' means we prune incomplete programs with the cost model during the sampling process and do not use fine-tuning.
``No fine-tuning'' is based on ``\Ansor (ours)'' but disables fine-tuning and only relies on random sampling.
``Limited space'' is also based on ``\Ansor (ours)'' but limits the search space to make it similar to the space in existing manual templates (\textit{e.g.}, limit tiling level, innermost tile sizes, and computation location).
As demonstrated by \autoref{fig:cpu-op-ablation}, dropping either the large search space or efficient fine-tuning decreases the final performance significantly. The aggressive early pruning in ``Beam search'' throws away incomplete programs with good final performance due to inaccurate estimation.

\subsection{Subgraph Benchmark}
\label{subsec:subgraph-benchmark}

\begin{figure}[t!]
	\centering
	\includegraphics[width=\columnwidth]{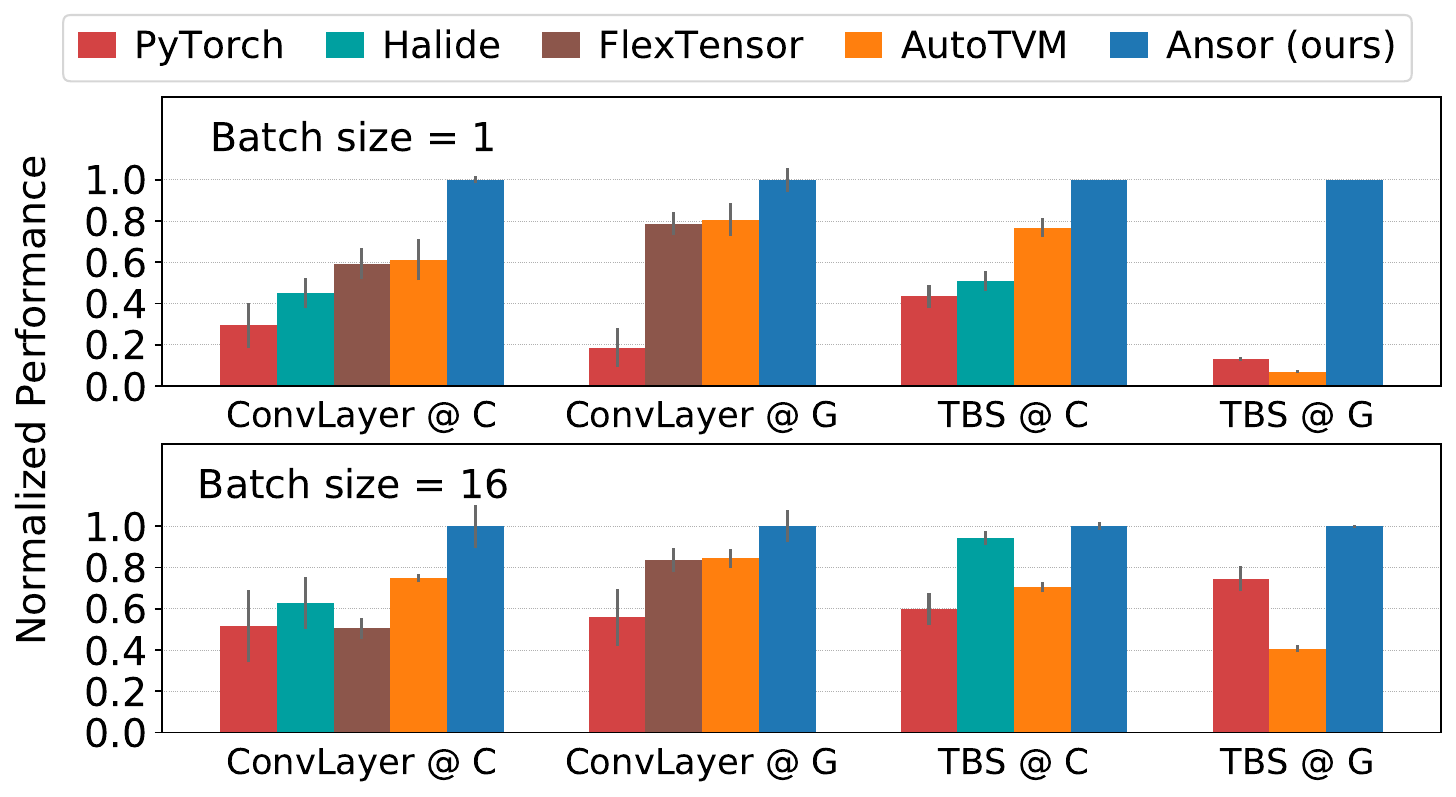}
	\precap
	\caption{Subgraph performance benchmark on a 20-core Intel-Platinum-8269CY and an NVIDIA V100. "@C" denotes CPU results and "@G" denotes GPU results. The y-axis is the throughput normalized to the best throughput for each subgraph.}
	\postcap
	\label{fig:subgraph}
\end{figure}

We perform the subgraph benchmark on two common subgraphs in DNNs.
The ``ConvLayer'' is a subgraph consisting of 2D convolution, batch normalization \cite{ioffe2015batch}, and ReLU activation, which is a common pattern in convolutional neural networks.
The ``TBS'' is a subgraph consisting of two matrix transposes, one batch matrix multiplication, and a softmax, which is a pattern in the multi-head attention \cite{vaswani2017attention} in language models.
Similar to the single operator benchmark (\autoref{subsec:single-op-benchmark}), we select four different shape configurations and two batch sizes, run auto-tuning with up to $1,000$ measurement trails per test case, and report the normalized performance.
We use the same set of baseline frameworks and run the benchmark on the Intel CPU and the NVIDIA GPU.
We do not report the performance of Halide auto-scheduler on GPU because as of writing the paper its GPU support is still in an experimental stage. FlexTensor fails to run on complicated subgraphs like ``TBS''.

\autoref{fig:subgraph} shows that \Ansor outperforms manual libraries and other search frameworks by $1.1-14.2\times$.
\Ansor can generate high-performance programs consistently for these subgraphs on both platforms.
FlexTensor performs well for single operators but shows less advantage for subgraphs because it lacks the support of operator fusion.

\subsection{End-to-End Network Benchmark}
\label{subsec:network-benchmark}

\textbf{Workloads.}
We benchmark the end-to-end inference execution time of several DNNs, which include ResNet-50~\cite{he2016deep} and MobileNet-V2~\cite{sandler2018mobilenetv2} for image classification, 3D-ResNet-18~\cite{hara3dcnns} for action recognition, DCGAN~\cite{radford2015unsupervised} generator for image generation, and BERT~\cite{devlin2018bert} for language understanding.
We benchmark these DNNs on three hardware platforms. For the server-class Intel CPU and NVIDIA GPU, we report the results for batch size 1 and batch size 16. For the ARM CPU in the edge device, real-time feedback is typically desired, so we only report the results for batch size 1.

\textbf{Baselines and Settings.}
We include PyTorch (v1.5 with torch script), TensorFlow (v2.0 with graph mode), TensorRT (v6.0 with TensorFlow integration)~\cite{nvidia2017tensorrt}, TensorFlow Lite (V2.0), and AutoTVM as baseline frameworks. We do not include Halide auto-scheduler or FlexTensor because they lack the support of widely-used deep learning model formats (\textit{e.g.}, ONNX, TensorFlow PB) and high-level graph optimizations.
As a result, we expect that the end-to-end execution time they can achieve will be the sum of the latency of all subgraphs in a DNN.
In contract, AutoTVM can optimize a whole DNN with its manual templates and various graph-level optimizations (\textit{e.g.}, graph-level layout search~\cite{liu2019optimizing}, graph-level constant folding~\cite{roesch2019relay}) which improve the performance significantly.
\Ansor also performs layout rewrite as described in \autoref{subsec:random-sampling}.
We let both AutoTVM and \Ansor run auto-tuning until they use to $1000 \times n$ measurement trials on each DNN, where $n$ is the number of subgraphs in the DNN. This is typically enough for them to converge.
We set the objective of the task scheduler as minimizing the total latency of one DNN and generate programs for these networks one by one.
On the other hand, PyTorch, TensorFlow, TensorRT, and TensorFlow Lite are all backed by static kernel libraries (MKL-DNN on Intel CPU, CuDNN on NVIDIA GPU, and Eigen on ARM CPU) and do not need auto-tuning. We enable AVX-512 for all frameworks on the Intel CPU in this network benchmark.

\begin{figure}[t]
	\centering
	\subfloat[Intel CPU]{
		\includegraphics[width=\columnwidth]{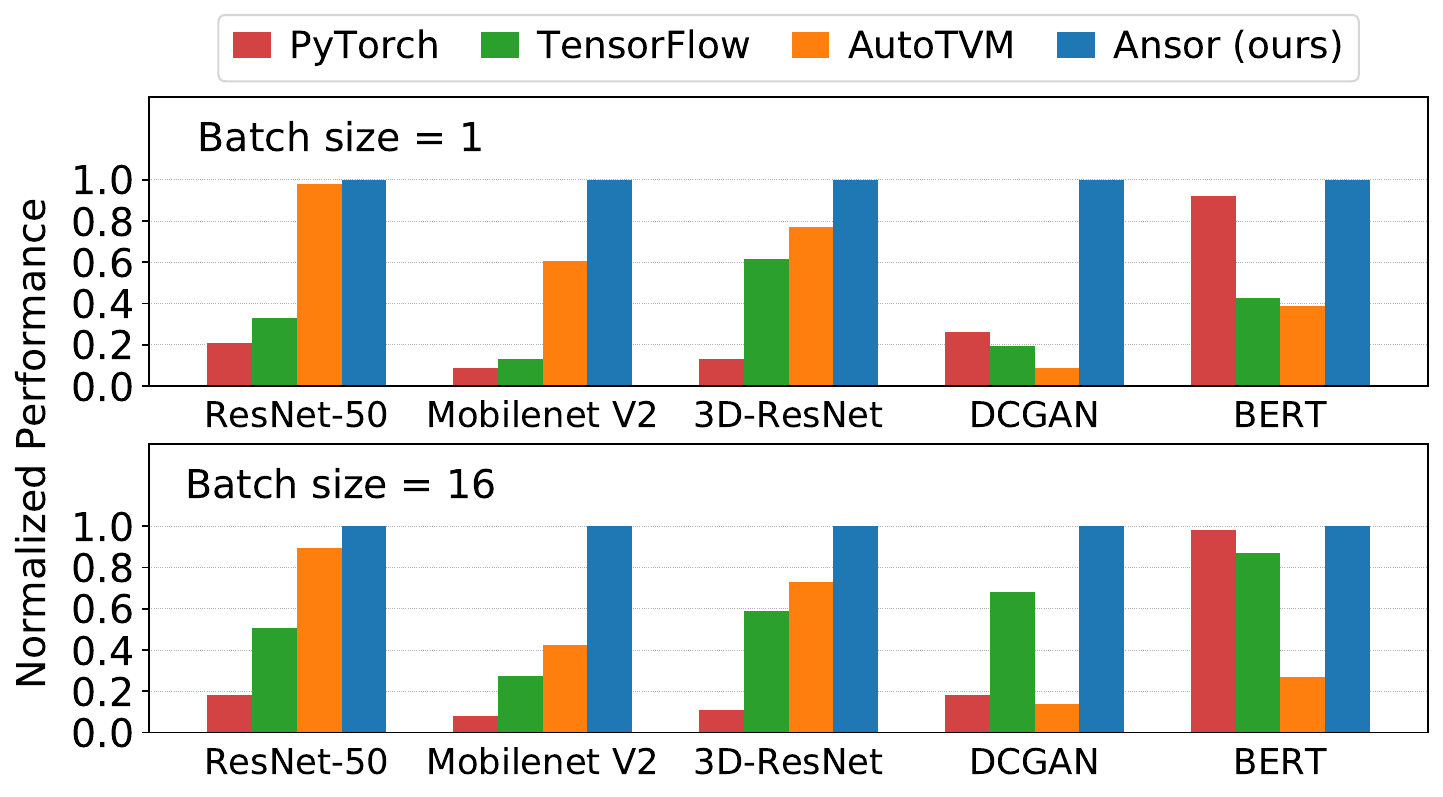}
		\label{fig:intel-cpu-network}
	}
	
	\subfloat[NVIDIA GPU]{
		\includegraphics[width=\columnwidth]{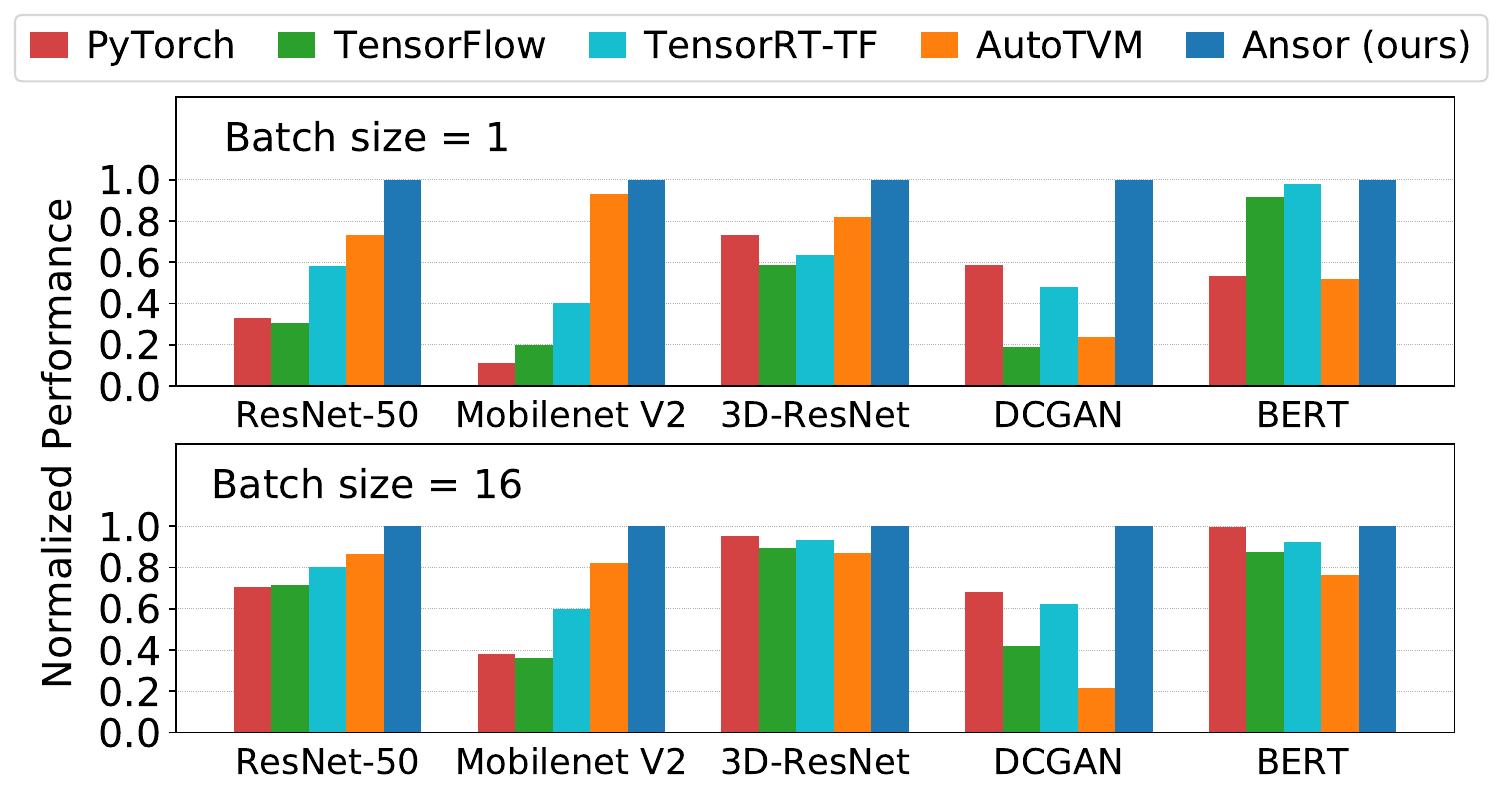}
		\label{fig:nvidia-gpu-metwork}
	}
	
	\subfloat[ARM CPU]{
		\includegraphics[width=\columnwidth]{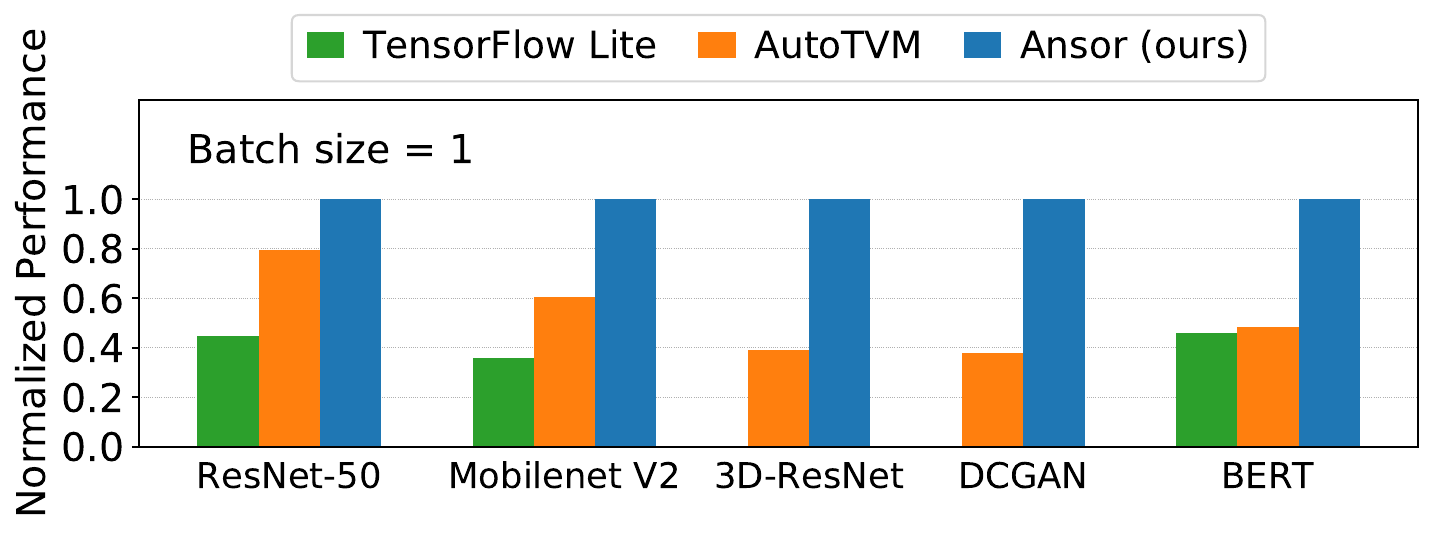}
		\label{fig:arm-cpu-network}
	}
	\precap
	\caption{Network inference performance benchmark on three hardware platforms. The y-axis is the throughput relative to the best throughput for each network.}
	\label{fig:network-eval} 
	\postcap
\end{figure}

\textbf{Results.}
\autoref{fig:network-eval} shows the results on the Intel CPU, NVIDIA GPU and ARM CPU \footnote{3D-ResNet and DCGAN are not yet supported by TensorFlow Lite on the ARM CPU.}.
Overall, \Ansor performs the best or equally the best in all cases.
Compared with search-based AutoTVM, \Ansor matches or outperforms it in all cases with $1.0-21.8 \times$ speedup.
Compared with the best alternative, \Ansor improves the execution performance of DNNs on the Intel CPU, ARM CPU, and NVIDIA GPU by up to  $3.8\times$, $2.6\times$, and $1.7 \times$, respectively. The reason for the significant speedup on DCGAN is that DCGAN mainly consists of transposed 2D convolution (T2D), which can be well optimized by \Ansor, as shown and explained in the single operator benchmark (\autoref{subsec:single-op-benchmark}).
AutoTVM performs very well for ResNet-50 on the Intel CPU thanks to its highly-optimized templates for 2D convolution and global layout search~\cite{liu2019optimizing}. \Ansor does not run a global layout search but does rewrite the layout of weight tensors as described in \autoref{subsec:random-sampling}. \Ansor uses more levels of tiling so it packs weight tensors into more levels. The layout rewrite brings about 40\% improvement to ResNet-50 in \Ansor.
Compared with vendor-specific static libraries, \Ansor has more advantages on uncommon shapes and small batch sizes, because it is not easy to manually optimize for these cases.

\begin{figure}[t!]
	\centering
	\includegraphics[width=\columnwidth]{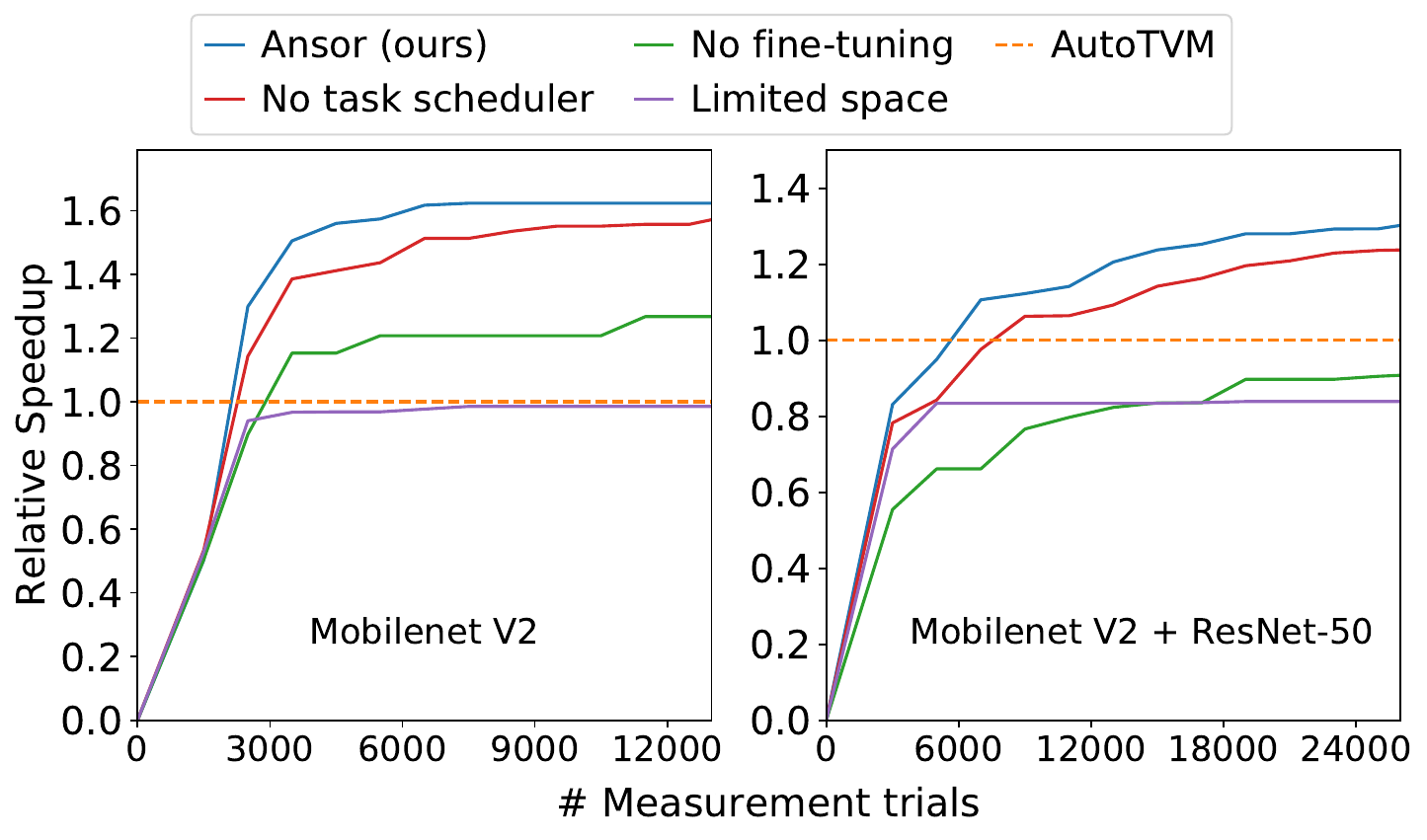}
	\precap
	\caption{Network performance auto-tuning curve. The y-axis is the speedup relative to AutoTVM.}
	\postcap \vskip -0.5em
	\label{fig:network-tuning-curve}
\end{figure}

\textbf{Ablation study.} We run variants of \Ansor on two test cases in \autoref{fig:network-tuning-curve}. In the left figure, we run four variants of \Ansor to generate programs for a single mobilenet-V2. In the right figure, we run these variants for both mobilenet-V2 and ResNet-50.
We set the objective function of the task scheduler to be the geometric mean of speedup against AutoTVM.
As shown in \autoref{fig:network-tuning-curve}, ``No task scheduler'' means we use a round-robin strategy to allocate equal time resources to all subgraphs. ``Limited space'' is based on ``\Ansor (ours)'' but limits the search space. ``No fine-tuning'' is also based on ``\Ansor (ours)'' but disables fine-tuning and relies on random sampling only.
As can be seen in \autoref{fig:network-tuning-curve}, ``Limited space'' performs the worst in terms of the final achieved performance, proving that the best programs are not included in the limited space.
The final achieved performance can be improved by enlarging the search space, as depicted in ``No fine-tuning''.
However, in the right figure, randomly assigning tile sizes and annotations still cannot beat AutoTVM in the given time budget.
After enabling fine-tuning, ``No task scheduler'' outperforms AutoTVM in both cases.
Finally, ``\Ansor (ours)'' employs the task scheduler to prioritize performance bottlenecks (\textit{e.g.}, subgraphs contain 3x3 convolution), so it performs the best in both search efficiency and the final achieved performance.

\subsection{Search Time}

\Ansor searches efficiently and can outperform or match AutoTVM with less search time.
\Ansor slices the time and utilizes the task scheduler to simultaneously optimize all subgraphs together.
In contrast, AutoTVM and other systems do not have a task scheduler, so they generate programs for all subgraphs one by one with a predefined budget of measurement trials for each subgraph.
\Ansor saves the search time by prioritizing important subgraphs, while AutoTVM spends predefined time budget on every subgraph, which may be a waste on the unimportant subgraphs.

\autoref{table:search-time} shows the search time required for \Ansor to match the performance of AutoTVM on the Intel CPU network benchmark (\autoref{subsec:network-benchmark}). We list the search time in two metrics: number of measurements and wall-clock time. ``Number of measurements'' is a metric agnostic to the implementation of measurement and the overhead of search algorithm, while ``Wall-clock time'' takes these factors into account.
As shown in the table, \Ansor can match the performance of AutoTVM with an order of magnitude less search time.
In \autorefsuffix{table:search-time}{a} the saving in search time comes from the task scheduler, efficient fine-tuning, and comprehensive coverage of effective optimizations.
In \autorefsuffix{table:search-time}{b}, \Ansor shows more time-saving in wall-clock time. This is because \Ansor does not introduce much search overhead and has a better implementation of the measurement (on the Intel CPU, \Ansor can get accurate measurement results with fewer repetitions by explicitly flushing the cache for some tensors).
On other backends, \Ansor can match the performance of AutoTVM with a similar saving in search time.

Typically, it takes several hours for \Ansor to generate fully-optimized programs for a DNN on a single machine. This is acceptable for inference applications because it is a one-shot effort before deployment. In addition, the whole architecture of \Ansor can be parallelized very easily.

\begin{table}[t]
	\centering
	\begin{tabular}{llll}
		\hline
		& AutoTVM & Ansor & Time-saving \\
		\hline
		ResNet-50    & 21,220   & 6,403 & 3.3 $\times$   \\
		Mobilenet-V2 & 31,272   & 1,892 & 16.5 $\times$  \\
		3D-ResNet    & 5,158    & 1,927 & 2.7 $\times$   \\
		DCGAN        & 3,003    & 298   & 10.1 $\times$  \\
		BERT         & 6,220    & 496   & 12.5 $\times$  \\
		\hline
	\end{tabular}
    \vskip -0.8 em
	\caption*{(a) The number of measurements.}

	\vskip 0.8em
	\centering
	\begin{tabular}{llll}
		\hline
		             & AutoTVM & \Ansor & Time-saving       \\ \hline
		ResNet-50    & 39,250  & 4,540  & 8.6 $\times$ \\
		Mobilenet-V2 & 58,468  & 660    & 88.6 $\times$ \\
		3D-ResNet    & 7,594   & 2,296  & 3.3  $\times$ \\
		DCGAN        & 4,914   & 420    & 11.7 $\times$ \\
		BERT         & 12,007  & 266    & 45.1 $\times$ \\ \hline
	\end{tabular}
    \vskip -0.8 em
	\caption*{(b) Wall-clock time (seconds) }
	
	\caption{The number of measurements and wall-clock time used for \Ansor to match the performance of AutoTVM on the Intel CPU (batch size=1).}
	\label{table:search-time}

\end{table}

\subsection{Cost Model Evaluation}
In this subsection, we evaluate the prediction quality of the learned cost model. 
We use 25,000 programs measured during tuning ResNet-50 on the Intel CPU as the data set.
We randomly pick 20,000 programs as the training set and use the remaining 5,000 programs as the test set.
We train the cost model and let it make predictions for the test set.

\begin{figure}[t!]
	\centering
	\includegraphics[width=\columnwidth]{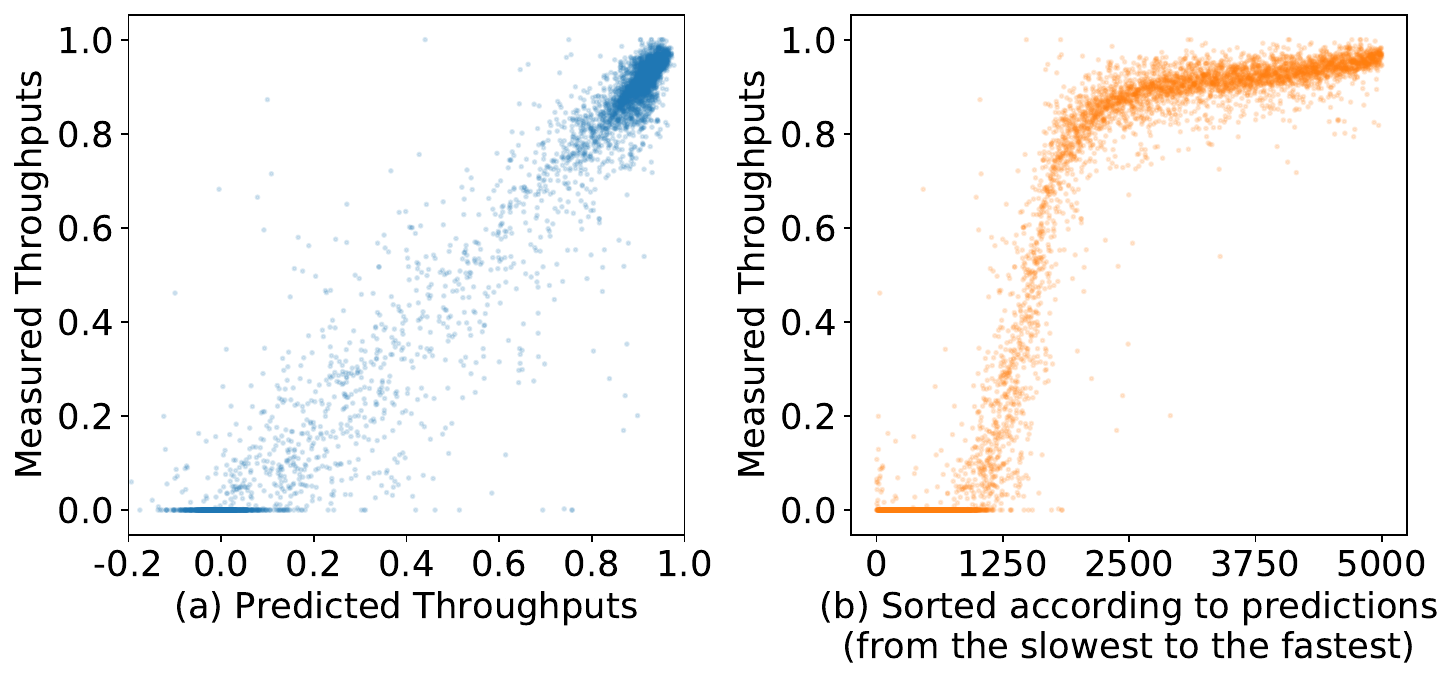}
	\precap
	\caption{Measured throughputs vs. predicted throughputs.}
	\postcap \vskip -0.5em
	\label{fig:cost-model-eval}
\end{figure}

\autoref{fig:cost-model-eval} plots the predicted throughputs vs. measured throughputs.
The measured throughputs are normalized to the best performing programs in the test set.
The predicted throughputs are the output of the model, so they can be negative.
In \autorefsuffix{fig:cost-model-eval}{a}, the points scatter around the diagonal line, meaning that the model makes accurate predictions.
The distribution is not uniform because the data set is collected during the search. Good programs have a higher probability to be chosen for measurements, so most of the programs are in the top right corner.
The points with measured throughput 0.0 are programs that are invalid or killed due to timeout during measurements.
In \autorefsuffix{fig:cost-model-eval}{b}, we sort the 5000 points according to the predictions from the slowest to the fastest, and use the relative ranking as x-axis.
So the points are distributed uniformly over x-axis. It shows the distribution of performance of the explored programs better.

The model archives  0.079 RMSE, 0.958 $R^2$ correlation, 0.851 pairwise comparison accuracy, and 0.624 recall@30 of top-30 programs (see the definition at \autoref{fn:top-k-recall-definition}) on the test set.

\section{Related Work}
\label{sec:related-work}

\textbf{Automatic tensor program generation based on scheduling languages.}
Halide \cite{ragan2013halide} introduces a scheduling language that can describe loop optimization primitives. This language is suitable for both manual optimization and automatic search. Halide has three versions of auto-scheduler based on different techniques \cite{mullapudi2016automatically,li2018differentiable,adams2019learning}. The latest one with beam search and learned cost model performs the best among them, which is also used in our evaluation.
TVM \cite{chen2018tvm} utilizes a similar scheduling language and includes a template-guided search framework
AutoTVM \cite{chen2018learning}.
FlexTensor\cite{zheng2020flextensor} proposes general templates that can target a set of operators,
but its templates are designed for single operators. It is hard to use these templates for optimizations involving multiple operators  (e.g., operator fusion).
A concurrent work ProTuner \cite{haj2020protuner} uses Monte Carlo tree search to solve the inaccurate estimation problem in Halide auto-scheduler. ProTuner mainly targets image processing workloads, while \Ansor targets deep learning workloads and introduces new search space and other optimizations. 

\textbf{Polyhedral compilation models.} The polyhedral compilation model \cite{bondhugula2008practical,verdoolaege2013polyhedral,verdoolaege2016presburger} formulates the optimization of programs as an integer linear programming (ILP) problem.
It optimizes a program with affine loop transformation that minimizes the data reuse distance between dependent statements. 
Tiramisu \cite{baghdadi2019tiramisu} and TensorComprehensions \cite{vasilache2018tensor} are two polyhedral compilers that also target the deep learning domain.
Tiramisu provides a scheduling language similar to the Halide language, and it needs manual scheduling.
TensorComprehensions can search for GPU code automatically,
but it is not yet meant to be used for compute-bounded problems \cite{chen2018tvm}. It cannot outperform TVM on operators like conv2d and matmul \cite{chen2018tvm, tillet2019triton}. 
This is because of the lack of certain optimizations \cite{vasilache2019next} and the inaccurate implicit cost model in the polyhedral formulation.

\textbf{Graph-level optimization for deep learning.} Graph-level optimizations treat an operator in the computational graph as a basic unit and perform optimization at graph level without changing the internal implementations of operators.
The common optimizations at graph level include layout optimizations \cite{liu2019optimizing}, operator fusion \cite{chen2018tvm, nvidia2017tensorrt, zheng2020fusionstitching}, constant folding ~\cite{roesch2019relay}, auto-batching ~\cite{looks2017deep}, automatic generation of graph substitution ~\cite{jia2019taso} and so forth. The graph-level optimizations are typically complementary to operator-level optimizations. Graph-level optimizations can also benefit from high-performance implementations of operators.
For example, general operator fusion relies on the code generation ability of \Ansor. We leave the joint optimization of \Ansor and more graph-level optimization as future work.

\textbf{Search-based compilation and auto-tuning.}
Search based compilation and auto-tuning have already shown their effectiveness in domains other than deep learning.
Stock \cite{schkufza2013stochastic} is a super-optimizer based on random search.
Stock searches for loop-free hardware instruction sequences, while \Ansor generates tensor programs with nests of loops.
OpenTuner \cite{ansel2014opentuner} is a general framework for program auto-tuning based on multi-armed bandit approaches.
OpenTuner relies on user-specified search space, while \Ansor constructs the search space automatically.
Traditional high-performance libraries such as ATLAS\cite{whaley1998automatically} and FFTW \cite{frigo1998fftw} also utilize auto-tuning.
More recent works NeuroVectorizer~\cite{haj2020neurovectorizer} and AutoPhase~\cite{huang2019autophase,haj2020autophase} use deep reinforcement learning to automatically vectorize programs and optimize the compiler phase ordering.

\section{Limitations and Future work}
\label{sec:limitation}
One of \Ansor's limitations is that \Ansor cannot optimize graphs with dynamic shapes~\cite{shen2020nimble}.
\Ansor requires the shapes in the computational graph to be static and known in advance
to do analysis, construct the search space, and perform measurements. How to generate programs for symbolic or dynamic shape is an interesting future direction.
Another limitation is that \Ansor only supports dense operators. 
To support sparse operators (\textit{e.g.}, SpMM) that are commonly used in sparse neural networks~\cite{gale2019state} and graph neural networks~\cite{hu2020featgraph}, we expect that a large portion of \Ansor can still be reused, but we need to redesign the search space.
Lastly, \Ansor only performs program optimizations at a high level but relies on other code generators (\textit{e.g.}, LLVM and NVCC) to do platform-dependent optimizations ({\textit{e.g.}, instruction selection}). \Ansor comes short of utilizing the special instructions, such as Intel VNNI, NVIDIA Tensor Core, and ARM Dot for mixed-precision and low-precision operators, which are not handled well by the off-the-shelf code generators currently.

\section{Conclusion}
\label{sec:conclusion}

We propose \Ansor, an automated search framework that generates high-performance tensor programs for deep neural networks. 
By efficiently exploring a large search space and prioritizing performance bottlenecks,
\Ansor finds high-performance programs that are outside the search space of existing approaches.
\Ansor outperforms existing manual libraries and search-based frameworks on a diverse set of neural networks and hardware platforms by up to $3.8 \times$.
By automatically searching for better programs, we hope that \Ansor will help bridge the gap between the increasing demand in computing power and limited hardware performance.
\Ansor is integrated into the Apache TVM open-source project \footnote{\url{https://tvm.apache.org/}}.

\section{Acknowledgement}
\label{sec:acknowledgement}

We would like to thank Weizhao Xian, Tianqi Chen, Frank Luan, anonymous reviewers, and our shepherd, Derek Murray, for their insightful feedback.
In addition to NSF CISE Expeditions Award CCF-1730628, this research is supported by gifts from Alibaba Group, Amazon Web Services, Ant Group, CapitalOne, Ericsson, Facebook, Futurewei, Google, Intel, Microsoft, Nvidia, Scotiabank, Splunk, and VMware.

\bibliographystyle{plain}
\bibliography{ansor.bib}

\begin{thebibliography}{10}

\bibitem{abadi2016tensorflow}
Mart{\'\i}n Abadi, Paul Barham, Jianmin Chen, Zhifeng Chen, Andy Davis, Jeffrey
  Dean, Matthieu Devin, Sanjay Ghemawat, Geoffrey Irving, Michael Isard, et~al.
\newblock Tensorflow: a system for large-scale machine learning.
\newblock In {\em 12th USENIX Symposium on Operating Systems Design and
  Implementation (OSDI 16)}, pages 265--283, 2016.

\bibitem{adams2019learning}
Andrew Adams, Karima Ma, Luke Anderson, Riyadh Baghdadi, Tzu-Mao Li,
  Micha{\"e}l Gharbi, Benoit Steiner, Steven Johnson, Kayvon Fatahalian,
  Fr{\'e}do Durand, et~al.
\newblock Learning to optimize halide with tree search and random programs.
\newblock {\em ACM Transactions on Graphics (TOG)}, 38(4):1--12, 2019.

\bibitem{alhaija2017augmented}
Hassan~Abu Alhaija, Siva~Karthik Mustikovela, Lars Mescheder, Andreas Geiger,
  and Carsten Rother.
\newblock Augmented reality meets deep learning for car instance segmentation
  in urban scenes.
\newblock In {\em British machine vision conference}, volume~1, page~2, 2017.

\bibitem{ansel2014opentuner}
Jason Ansel, Shoaib Kamil, Kalyan Veeramachaneni, Jonathan Ragan-Kelley,
  Jeffrey Bosboom, Una-May O'Reilly, and Saman Amarasinghe.
\newblock Opentuner: an extensible framework for program autotuning.
\newblock In {\em Proceedings of the 23rd international conference on Parallel
  architectures and compilation}, pages 303--316, 2014.

\bibitem{baghdadi2019tiramisu}
Riyadh Baghdadi, Jessica Ray, Malek~Ben Romdhane, Emanuele Del~Sozzo,
  Abdurrahman Akkas, Yunming Zhang, Patricia Suriana, Shoaib Kamil, and Saman
  Amarasinghe.
\newblock Tiramisu: a polyhedral compiler for expressing fast and portable
  code.
\newblock In {\em 2019 IEEE/ACM International Symposium on Code Generation and
  Optimization (CGO)}, pages 193--205. IEEE, 2019.

\bibitem{bai2019onnx}
Junjie Bai, Fang Lu, Ke~Zhang, et~al.
\newblock Onnx: open neural network exchange, 2019.

\bibitem{barham2019machine}
Paul Barham and Michael Isard.
\newblock Machine learning systems are stuck in a rut.
\newblock In {\em Proceedings of the Workshop on Hot Topics in Operating
  Systems}, pages 177--183, 2019.

\bibitem{bondhugula2008practical}
Uday Bondhugula, Albert Hartono, Jagannathan Ramanujam, and Ponnuswamy
  Sadayappan.
\newblock A practical automatic polyhedral parallelizer and locality optimizer.
\newblock In {\em Proceedings of the 29th ACM SIGPLAN Conference on Programming
  Language Design and Implementation}, pages 101--113, 2008.

\bibitem{chen2016xgboost}
Tianqi Chen and Carlos Guestrin.
\newblock Xgboost: a scalable tree boosting system.
\newblock In {\em Proceedings of the 22nd acm sigkdd international conference
  on knowledge discovery and data mining}, pages 785--794, 2016.

\bibitem{chen2015mxnet}
Tianqi Chen, Mu~Li, Yutian Li, Min Lin, Naiyan Wang, Minjie Wang, Tianjun Xiao,
  Bing Xu, Chiyuan Zhang, and Zheng Zhang.
\newblock Mxnet: a flexible and efficient machine learning library for
  heterogeneous distributed systems.
\newblock {\em arXiv preprint arXiv:1512.01274}, 2015.

\bibitem{chen2018tvm}
Tianqi Chen, Thierry Moreau, Ziheng Jiang, Lianmin Zheng, Eddie Yan, Haichen
  Shen, Meghan Cowan, Leyuan Wang, Yuwei Hu, Luis Ceze, et~al.
\newblock Tvm: an automated end-to-end optimizing compiler for deep learning.
\newblock In {\em 13th USENIX Symposium on Operating Systems Design and
  Implementation (OSDI 18)}, pages 578--594, 2018.

\bibitem{chen2018learning}
Tianqi Chen, Lianmin Zheng, Eddie Yan, Ziheng Jiang, Thierry Moreau, Luis Ceze,
  Carlos Guestrin, and Arvind Krishnamurthy.
\newblock Learning to optimize tensor programs.
\newblock In {\em Advances in Neural Information Processing Systems}, pages
  3389--3400, 2018.

\bibitem{chetlur2014cudnn}
Sharan Chetlur, Cliff Woolley, Philippe Vandermersch, Jonathan Cohen, John
  Tran, Bryan Catanzaro, and Evan Shelhamer.
\newblock cudnn: efficient primitives for deep learning.
\newblock {\em arXiv preprint arXiv:1410.0759}, 2014.

\bibitem{cordts2016cityscapes}
Marius Cordts, Mohamed Omran, Sebastian Ramos, Timo Rehfeld, Markus Enzweiler,
  Rodrigo Benenson, Uwe Franke, Stefan Roth, and Bernt Schiele.
\newblock The cityscapes dataset for semantic urban scene understanding.
\newblock In {\em Proceedings of the IEEE conference on computer vision and
  pattern recognition}, pages 3213--3223, 2016.

\bibitem{devlin2018bert}
Jacob Devlin, Ming-Wei Chang, Kenton Lee, and Kristina Toutanova.
\newblock Bert: pre-training of deep bidirectional transformers for language
  understanding.
\newblock {\em arXiv preprint arXiv:1810.04805}, 2018.

\bibitem{frigo1998fftw}
Matteo Frigo and Steven~G Johnson.
\newblock Fftw: an adaptive software architecture for the fft.
\newblock In {\em Proceedings of the 1998 IEEE International Conference on
  Acoustics, Speech and Signal Processing, ICASSP'98 (Cat. No. 98CH36181)},
  volume~3, pages 1381--1384. IEEE, 1998.

\bibitem{gale2019state}
Trevor Gale, Erich Elsen, and Sara Hooker.
\newblock The state of sparsity in deep neural networks.
\newblock {\em arXiv preprint arXiv:1902.09574}, 2019.

\bibitem{haj2020neurovectorizer}
Ameer Haj-Ali, Nesreen~K Ahmed, Ted Willke, Yakun~Sophia Shao, Krste Asanovic,
  and Ion Stoica.
\newblock Neurovectorizer: end-to-end vectorization with deep reinforcement
  learning.
\newblock In {\em Proceedings of the 18th ACM/IEEE International Symposium on
  Code Generation and Optimization}, pages 242--255, 2020.

\bibitem{haj2020protuner}
Ameer Haj-Ali, Hasan Genc, Qijing Huang, William Moses, John Wawrzynek, Krste
  Asanovi{\'c}, and Ion Stoica.
\newblock Protuner: tuning programs with monte carlo tree search.
\newblock {\em arXiv preprint arXiv:2005.13685}, 2020.

\bibitem{haj2020autophase}
Ameer Haj-Ali, Qijing Huang, William Moses, John Xiang, John Wawrzynek, Krste
  Asanovic, and Ion Stoica.
\newblock Autophase: juggling hls phase orderings in random forests with deep
  reinforcement learning.
\newblock In {\em Third Conference on Machine Learning and Systems (ML-Sys)},
  2020.

\bibitem{hara3dcnns}
Kensho Hara, Hirokatsu Kataoka, and Yutaka Satoh.
\newblock Can spatiotemporal 3d cnns retrace the history of 2d cnns and
  imagenet?
\newblock In {\em Proceedings of the IEEE Conference on Computer Vision and
  Pattern Recognition (CVPR)}, pages 6546--6555, 2018.

\bibitem{he2016deep}
Kaiming He, Xiangyu Zhang, Shaoqing Ren, and Jian Sun.
\newblock Deep residual learning for image recognition.
\newblock In {\em Proceedings of the IEEE conference on computer vision and
  pattern recognition}, pages 770--778, 2016.

\bibitem{hinton2018matrix}
Geoffrey~E Hinton, Sara Sabour, and Nicholas Frosst.
\newblock Matrix capsules with em routing.
\newblock 2018.

\bibitem{howard2017mobilenets}
Andrew~G Howard, Menglong Zhu, Bo~Chen, Dmitry Kalenichenko, Weijun Wang,
  Tobias Weyand, Marco Andreetto, and Hartwig Adam.
\newblock Mobilenets: efficient convolutional neural networks for mobile vision
  applications.
\newblock {\em arXiv preprint arXiv:1704.04861}, 2017.

\bibitem{hu2020featgraph}
Yuwei Hu, Zihao Ye, Minjie Wang, Jiali Yu, Da~Zheng, Mu~Li, Zheng Zhang, Zhiru
  Zhang, and Yida Wang.
\newblock Featgraph: A flexible and efficient backend for graph neural network
  systems.
\newblock {\em arXiv preprint arXiv:2008.11359}, 2020.

\bibitem{huang2019autophase}
Qijing Huang, Ameer Haj-Ali, William Moses, John Xiang, Ion Stoica, Krste
  Asanovic, and John Wawrzynek.
\newblock Autophase: compiler phase-ordering for hls with deep reinforcement
  learning.
\newblock In {\em 2019 IEEE 27th Annual International Symposium on
  Field-Programmable Custom Computing Machines (FCCM)}, pages 308--308. IEEE,
  2019.

\bibitem{intel2017mkldnn}
Intel.
\newblock Intel® math kernel library for deep learning networks, 2017.

\bibitem{ioffe2015batch}
Sergey Ioffe and Christian Szegedy.
\newblock Batch normalization: accelerating deep network training by reducing
  internal covariate shift.
\newblock {\em arXiv preprint arXiv:1502.03167}, 2015.

\bibitem{jia2019taso}
Zhihao Jia, Oded Padon, James Thomas, Todd Warszawski, Matei Zaharia, and Alex
  Aiken.
\newblock Taso: optimizing deep learning computation with automatic generation
  of graph substitutions.
\newblock In {\em Proceedings of the 27th ACM Symposium on Operating Systems
  Principles}, pages 47--62, 2019.

\bibitem{lavin2016fast}
Andrew Lavin and Scott Gray.
\newblock Fast algorithms for convolutional neural networks.
\newblock In {\em Proceedings of the IEEE Conference on Computer Vision and
  Pattern Recognition}, pages 4013--4021, 2016.

\bibitem{li2018differentiable}
Tzu-Mao Li, Micha{\"e}l Gharbi, Andrew Adams, Fr{\'e}do Durand, and Jonathan
  Ragan-Kelley.
\newblock Differentiable programming for image processing and deep learning in
  halide.
\newblock {\em ACM Transactions on Graphics (TOG)}, 37(4):139, 2018.

\bibitem{liu2019optimizing}
Yizhi Liu, Yao Wang, Ruofei Yu, Mu~Li, Vin Sharma, and Yida Wang.
\newblock Optimizing cnn model inference on cpus.
\newblock In {\em 2019 USENIX Annual Technical Conference (USENIX ATC 19)},
  pages 1025--1040, 2019.

\bibitem{looks2017deep}
Moshe Looks, Marcello Herreshoff, DeLesley Hutchins, and Peter Norvig.
\newblock Deep learning with dynamic computation graphs.
\newblock {\em arXiv preprint arXiv:1702.02181}, 2017.

\bibitem{medress1977speech}
Mark~F. Medress, Franklin~S Cooper, Jim~W. Forgie, CC~Green, Dennis~H. Klatt,
  Michael~H. O'Malley, Edward~P Neuburg, Allen Newell, DR~Reddy, B~Ritea,
  et~al.
\newblock Speech understanding systems: report of a steering committee.
\newblock {\em Artificial Intelligence}, 9(3):307--316, 1977.

\bibitem{moreau2019hardware}
Thierry Moreau, Tianqi Chen, Luis Vega, Jared Roesch, Eddie Yan, Lianmin Zheng,
  Josh Fromm, Ziheng Jiang, Luis Ceze, Carlos Guestrin, et~al.
\newblock A hardware--software blueprint for flexible deep learning
  specialization.
\newblock {\em IEEE Micro}, 39(5):8--16, 2019.

\bibitem{mullapudi2016automatically}
Ravi~Teja Mullapudi, Andrew Adams, Dillon Sharlet, Jonathan Ragan-Kelley, and
  Kayvon Fatahalian.
\newblock Automatically scheduling halide image processing pipelines.
\newblock {\em ACM Transactions on Graphics (TOG)}, 35(4):83, 2016.

\bibitem{nvidia2017tensorcore}
Nvidia.
\newblock Nvidia tensor cores, 2017.

\bibitem{nvidia2017tensorrt}
Nvidia.
\newblock Nvidia tensorrt: programmable inference accelerator, 2017.

\bibitem{paszke2019pytorch}
Adam Paszke, Sam Gross, Francisco Massa, Adam Lerer, James Bradbury, Gregory
  Chanan, Trevor Killeen, Zeming Lin, Natalia Gimelshein, Luca Antiga, et~al.
\newblock Pytorch: an imperative style, high-performance deep learning library.
\newblock In {\em Advances in Neural Information Processing Systems}, pages
  8024--8035, 2019.

\bibitem{radford2015unsupervised}
Alec Radford, Luke Metz, and Soumith Chintala.
\newblock Unsupervised representation learning with deep convolutional
  generative adversarial networks.
\newblock {\em arXiv preprint arXiv:1511.06434}, 2015.

\bibitem{ragan2013halide}
Jonathan Ragan-Kelley, Connelly Barnes, Andrew Adams, Sylvain Paris, Fr{\'e}do
  Durand, and Saman Amarasinghe.
\newblock Halide: a language and compiler for optimizing parallelism, locality,
  and recomputation in image processing pipelines.
\newblock {\em Acm Sigplan Notices}, 48(6):519--530, 2013.

\bibitem{roesch2019relay}
Jared Roesch, Steven Lyubomirsky, Marisa Kirisame, Josh Pollock, Logan Weber,
  Ziheng Jiang, Tianqi Chen, Thierry Moreau, and Zachary Tatlock.
\newblock Relay: a high-level compiler for deep learning.
\newblock {\em arXiv preprint arXiv:1904.08368}, 2019.

\bibitem{sandler2018mobilenetv2}
Mark Sandler, Andrew Howard, Menglong Zhu, Andrey Zhmoginov, and Liang-Chieh
  Chen.
\newblock Mobilenetv2: inverted residuals and linear bottlenecks.
\newblock In {\em Proceedings of the IEEE conference on computer vision and
  pattern recognition}, pages 4510--4520, 2018.

\bibitem{schkufza2013stochastic}
Eric Schkufza, Rahul Sharma, and Alex Aiken.
\newblock Stochastic superoptimization.
\newblock {\em ACM SIGARCH Computer Architecture News}, 41(1):305--316, 2013.

\bibitem{shen2020nimble}
Haichen Shen, Jared Roesch, Zhi Chen, Wei Chen, Yong Wu, Mu~Li, Vin Sharma,
  Zachary Tatlock, and Yida Wang.
\newblock Nimble: Efficiently compiling dynamic neural networks for model
  inference.
\newblock {\em arXiv preprint arXiv:2006.03031}, 2020.

\bibitem{suriana2017parallel}
Patricia Suriana, Andrew Adams, and Shoaib Kamil.
\newblock Parallel associative reductions in halide.
\newblock In {\em 2017 IEEE/ACM International Symposium on Code Generation and
  Optimization (CGO)}, pages 281--291. IEEE, 2017.

\bibitem{sutton2018reinforcement}
Richard~S Sutton and Andrew~G Barto.
\newblock {\em Reinforcement learning: an introduction}.
\newblock MIT press, 2018.

\bibitem{tillet2019triton}
Philippe Tillet, HT~Kung, and David Cox.
\newblock Triton: an intermediate language and compiler for tiled neural
  network computations.
\newblock In {\em Proceedings of the 3rd ACM SIGPLAN International Workshop on
  Machine Learning and Programming Languages}, pages 10--19, 2019.

\bibitem{vasilache2018tensor}
Nicolas Vasilache, Oleksandr Zinenko, Theodoros Theodoridis, Priya Goyal,
  Zachary DeVito, William~S Moses, Sven Verdoolaege, Andrew Adams, and Albert
  Cohen.
\newblock Tensor comprehensions: framework-agnostic high-performance machine
  learning abstractions.
\newblock {\em arXiv preprint arXiv:1802.04730}, 2018.

\bibitem{vasilache2019next}
Nicolas Vasilache, Oleksandr Zinenko, Theodoros Theodoridis, Priya Goyal,
  Zachary Devito, William~S Moses, Sven Verdoolaege, Andrew Adams, and Albert
  Cohen.
\newblock The next 700 accelerated layers: from mathematical expressions of
  network computation graphs to accelerated gpu kernels, automatically.
\newblock {\em ACM Transactions on Architecture and Code Optimization (TACO)},
  16(4):1--26, 2019.

\bibitem{vaswani2017attention}
Ashish Vaswani, Noam Shazeer, Niki Parmar, Jakob Uszkoreit, Llion Jones,
  Aidan~N Gomez, {\L}ukasz Kaiser, and Illia Polosukhin.
\newblock Attention is all you need.
\newblock In {\em Advances in neural information processing systems}, pages
  5998--6008, 2017.

\bibitem{verdoolaege2016presburger}
Sven Verdoolaege.
\newblock Presburger formulas and polyhedral compilation.
\newblock 2016.

\bibitem{verdoolaege2013polyhedral}
Sven Verdoolaege, Juan Carlos~Juega, Albert Cohen, Jose Ignacio~Gomez,
  Christian Tenllado, and Francky Catthoor.
\newblock Polyhedral parallel code generation for cuda.
\newblock {\em ACM Transactions on Architecture and Code Optimization (TACO)},
  9(4):1--23, 2013.

\bibitem{vikhar2016evolutionary}
Pradnya~A Vikhar.
\newblock Evolutionary algorithms: a critical review and its future prospects.
\newblock In {\em 2016 International conference on global trends in signal
  processing, information computing and communication (ICGTSPICC)}, pages
  261--265. IEEE, 2016.

\bibitem{wang2019unified}
Leyuan Wang, Zhi Chen, Yizhi Liu, Yao Wang, Lianmin Zheng, Mu~Li, and Yida
  Wang.
\newblock A unified optimization approach for cnn model inference on integrated
  gpus.
\newblock In {\em Proceedings of the 48th International Conference on Parallel
  Processing}, pages 1--10, 2019.

\bibitem{whaley1998automatically}
R~Clinton Whaley and Jack~J Dongarra.
\newblock Automatically tuned linear algebra software.
\newblock In {\em SC'98: Proceedings of the 1998 ACM/IEEE conference on
  Supercomputing}, pages 38--38. IEEE, 1998.

\bibitem{yu2015multi}
Fisher Yu and Vladlen Koltun.
\newblock Multi-scale context aggregation by dilated convolutions.
\newblock {\em arXiv preprint arXiv:1511.07122}, 2015.

\bibitem{zheng2020flextensor}
Size Zheng, Yun Liang, Shuo Wang, Renze Chen, and Kaiwen Sheng.
\newblock Flextensor: an automatic schedule exploration and optimization
  framework for tensor computation on heterogeneous system.
\newblock In {\em Proceedings of the Twenty-Fifth International Conference on
  Architectural Support for Programming Languages and Operating Systems}, pages
  859--873, 2020.

\bibitem{zheng2020fusionstitching}
Zhen Zheng, Pengzhan Zhao, Guoping Long, Feiwen Zhu, Kai Zhu, Wenyi Zhao,
  Lansong Diao, Jun Yang, and Wei Lin.
\newblock Fusionstitching: boosting memory intensive computations for deep
  learning workloads.
\newblock {\em arXiv preprint arXiv:2009.10924}, 2020.

\end{thebibliography}

\appendix
\section{Gradient Approximation for the Task Scheduler}
\label{sec:gradient-approximation}
Now we show how to approximate the gradient for the objective function $f$. First, do the approximation $g_i(t) \approx g_i(t_i)$.
This means we assume the best cost of task $i$ depends only on the resource units spent on it. This may not be true because all tasks share a cost model. Different resource allocations lead to different collections of training data, which then leads to different cost models.
Here we make this approximation to continue derivation:

\begin{align*}
\frac{\partial f}{\partial t_i} &= \frac{\partial f}{\partial g_i} \frac{\partial g_i}{\partial t_i} \\
&\approx \frac{\partial f}{\partial g_i} (\alpha \frac{g_i(t_i) - g_i(t_i - \Delta t)}{\Delta t}  
+ (1-\alpha)  \frac{g_i(t_i + \Delta t) - g_i(t_i)}{\Delta t} ) \\
&\approx \frac{\partial f}{\partial g_i} (\alpha \frac{g_i(t_i) - g_i(t_i - \Delta t)}{\Delta t}  
+ (1-\alpha)  (g_i(t_i + 1) - g_i(t_i))) \\
\end{align*}

\noindent In this expression, $\Delta t$ is a small backward window size, $g_i(t_i)$ and $g_i(t_i - \Delta t)$ are known from the history allocations. But $g_i(t_i + 1)$ is unknown because we have not allocated $t_i + 1$ units of resource to this task. So we have to predict this value.
The parameter $\alpha$ controls the weight to trust the prediction.
We predict $g_i(t_i + 1)$ in two ways. First, we have an optimistic guess that if we spend extra $t_i$, we can decrease the latency of task $i$ to 0. This means $g_i(t_i + 1) \approx g_i(t_i) - \frac{g_i(t_i)}{t_i}$.
Second, if subgraphs are structurally similar, their latency is also similar per floating point operation.
Considering both factors, we have the following approximation:
$$g_i(t_i + 1) \approx \min(g_i(t_i) - \frac{g_i(t_i)}{t_i}, \beta \frac{C_i}{\max_{k\in N(i)} {V_k}}) $$
\noindent where $N(i)$ is the set of similar tasks of $i$, $C_i$ is the number of floating point operations in task $i$ and $V_k$ is the number of floating point operation per second we can achieve in task $k$. The parameter $\beta$ controls the weight to trust the prediction based on similarity.

\section{The List of Extracted Features}
\label{sec:extracted-features}
We extract the following features for one innermost non-loop statement in the context of a full tensor program. The features include categorical features and numerical features. We use one-hot encoding to encode category features. The length of a feature vector including all the listed features for one statement is $164$. We use the same set of features for both CPU and GPU.

\begin{itemize}
    \itemsep-0.2em
    \item \textbf{Numbers of float operations}. The numbers of addition, subtraction, division, modulo operation, comparison, intrinsic math function call (\textit{e.g,}. exp, sqrt) and other math function call respectively, with floating point operands.
    \item \textbf{Number of integer operations.} Similar to the above one, but for operations with integer operands.
    \item \textbf{Vectorization related features.} The length of the innermost vectorized loop. The type of vectorization position (InnerSpatial, MiddleSpatial, OuterSpatial, InnerReduce, MiddleReduce, OuterReduce, Mixed, None). The product of the lengths of all vectorized loops. The number of vectorized loops.
    \item \textbf{Unrolling related features.} Similar to the vectorization related features, but for unrolling.
    \item \textbf{Parallelization related features.} Similar to the vectorization related features, but for parallelization.
    \item \textbf{GPU thread binding related features.} The lengths of blockIdx.x, blockIdx.y, blockIdx.z, threadIdx.x, threadIdx.y, threadIdx.z and virtual threads \cite{chen2018tvm}.
    \item \textbf{Arithmetic intensity curve.} Arithmetic intensity is defined as  $\frac{\text{The number of floating point operations}}{\text{The number of bytes accessed}}$. We compute the arithmetic intensity for each loop level and draw a curve with linear interpolation. Then we sample 10 points from this curve.
    \item \textbf{Buffer Access Feature.}
    For each buffer this statement accesses, we extract features for it. While different statements can access different numbers of buffers, we perform feature extraction for at most five buffers. We pad zeros if a statement accesses less than five buffers and remove small buffers if a statement accesses more than five buffers.
    \begin{itemize}
        \item \textbf{Access type}. The type of access (read, write, read + write).
        \item \textbf{Bytes}. The total number of bytes accessed by this statement.
        \item \textbf{Unique bytes}. The total number of unique bytes accessed by this statement.
        \item \textbf{Lines}. The total number of cache lines accessed by this statement.
        \item \textbf{Unique lines}. The total number of unique cache lines accessed by this statement.
        \item \textbf{Reuse type}. The type of data reuse (LoopMultipleRead, SerialMultipleRead, NoReuse).
        \item \textbf{Reuse distance}.  The distance between data reuse in terms of number of for loop iterations and total accessed bytes.
        \item \textbf{Reuse counter}. The number of happening of data reuse.
        \item \textbf{Stride}.  The stride of access.
        \item \textbf{Accessed bytes divided by reuse}. We compute the following values: 
        $\frac{\text{Bytes}}{\text{Reuse counter}}$,
        $\frac{\text{Unique bytes}}{\text{Reuse counter}}$,
        $\frac{\text{Lines}}{\text{Reuse counter}}$,
        $\frac{\text{Unique lines}}{\text{Reuse counter}}$.
    \end{itemize}
    
    \item \textbf{Allocation related features}. The size of the allocated buffer for the output buffer of this statement. The number of allocations.
    \item \textbf{Other features.} The number of outer loops. The product of the lengths of outer loops. The value of the ``auto\_unroll\_max\_step''' specified by the pragma in outer loops.
\end{itemize}

\section{Shape Configurations in the Evaluation}
\label{sec:shape-eval}

\begin{itemize}
	\item C1D (1D Convolution). Format = (length, input channel, output channel, kernel size, stride, padding)
	\begin{itemize}[topsep=-0.1em, itemsep=-0.2em]
		\item (256, 64, 128, 3, 2, 1)
		\item (128, 128, 256, 1, 2, 0)
		\item (64, 256, 256, 5, 1, 2)
		\item (32, 512, 512, 3, 1, 1)
	\end{itemize}
	
	\item C2D (2D Convolution). Format = (height, width, input channel, output channel, kernel size, stride, padding)
	\begin{itemize}[topsep=-0.1em, itemsep=-0.2em]
		\item (224, 224, 3, 64, 7, 2, 3)
		\item (56, 56, 64, 64, 1, 1, 0)
		\item  (14, 14, 256, 256, 3, 1, 1)
		\item (7, 7, 512, 512, 3, 1, 1)
	\end{itemize}
	
	\item C3D (3D Convolution). Format = (depth, height, width, input channel, output channel, kernel size, stride, padding)
	\begin{itemize}[topsep=-0.1em, itemsep=-0.2em]
		\item (16, 224, 224, 3, 64, 7, 2, 3)
		\item (16, 56, 56, 64, 64, 1, 1, 0)
		\item  (16, 14, 14, 256, 256, 3, 1, 1)
		\item (16, 7, 7, 512, 512, 3, 1, 1)
	\end{itemize}
	
	\item GMM (Matrix Multiply). Format = (N, M, K)
	\begin{itemize}[topsep=-0.1em, itemsep=-0.2em]
		\item (128, 128, 128)
		\item (512, 32, 512)
		\item (512, 512, 512)
		\item (1024, 1024, 1024)
	\end{itemize}
	
	\item GRD (Group Convolution). Format = (height, width, input channel, output channel, kernel size, stride, padding, groups)
	\begin{itemize}[topsep=-0.1em, itemsep=-0.2em]
		\item (224, 224, 3, 64, 7, 2, 3, 4)
		\item (56, 56, 64, 64, 1, 1, 0, 4)
		\item  (14, 14, 256, 256, 3, 1, 1, 4)
		\item (7, 7, 512, 512, 3, 1, 1, 4)
	\end{itemize}
	
	\item DIL (Dilated Convolution). Format = (height, width, input channel, output channel, kernel size, stride, padding, dilation)
	\begin{itemize}[topsep=-0.1em, itemsep=-0.2em]
		\item (224, 224, 3, 64, 7, 2, 3, 2)
		\item (56, 56, 64, 64, 1, 1, 0, 2)
		\item  (14, 14, 256, 256, 3, 1, 1, 2)
		\item (7, 7, 512, 512, 3, 1, 1, 2)
	\end{itemize}
	
	\item DEP (Depthwise Convolution). Format = (height, width, channel, kernel size, stride, padding)
	\begin{itemize}[topsep=-0.1em, itemsep=-0.2em]
		\item (112, 112, 32,  3, 1, 1)
		\item (112, 112, 64,  3, 2, 1)
		\item (14,  14, 512, 3, 2, 1)
		\item (7,   7, 1024, 3, 1, 1)
	\end{itemize}
	
	\item T2D (Transposed 2D Convolution). Format = (height, width, input channel, output channel, kernel size, stride, padding)
	\begin{itemize}[topsep=-0.1em, itemsep=-0.2em]
		\item (4, 4, 512, 256, 4, 2, 1)
		\item (8, 8, 256, 128, 4, 2, 1)
		\item (16, 16, 128, 64, 4, 2, 1)
		\item (32, 32, 64, 3, 4, 2, 1)
	\end{itemize}
	
	\item CAP (Capsule 2D Convolution). Format = (height, width, input channel, output channel, kernel size, stride, padding, capsule size)
	\begin{itemize}[topsep=-0.1em, itemsep=-0.2em]
		\item (16, 16, 32, 32, 3, 2, 1, 4)
		\item (8,  8, 32, 32, 3, 1, 1, 4)
		\item (16, 16,  8, 16, 3, 2, 1, 4)
		\item (8,  8, 16, 16, 3, 1, 1, 4)
	\end{itemize}
	
	\item NRM (Matrix 2-Norm). Format = (N, M)
	\begin{itemize}[topsep=-0.1em, itemsep=-0.2em]
		\item (256, 256)
		\item (512, 512)
		\item (1024, 1024)
		\item (4096, 4096)
	\end{itemize}
	
	\item ConvLayer (Convolution Layer). Format = (height, width, input channel, output channel, kernel size, stride, padding)
	\begin{itemize}[topsep=-0.1em, itemsep=-0.2em]
		\item (224, 224, 3, 64, 7, 2, 3)
		\item (56, 56, 64, 64, 3, 2, 1)
		\item  (28, 28, 128, 256, 1, 2, 0)
		\item (7, 7, 512, 512, 3, 1, 1)
	\end{itemize}
	
	\item TBS (Transposed + BatchMatmul + Softmax in the multi-head attention). Format = (sequence length, number of heads, hidden dimension))
	\begin{itemize}[topsep=-0.1em, itemsep=-0.2em]
		\item (128, 12, 64)
		\item (128, 16, 64)
		\item (64,  12, 128)
		\item (128, 12, 128)
	\end{itemize}
\end{itemize}

\end{document}